\documentclass[]{interact}

\usepackage{graphicx}
\usepackage{epstopdf}
\DeclareGraphicsExtensions{.eps,.pdf}

\usepackage{ragged2e}

\usepackage{subfigure}

\usepackage{array}
\usepackage{mdwmath}
\usepackage{mdwtab}
\usepackage{eqparbox}
\usepackage{url}
\usepackage{amsmath}
\usepackage{float}
\usepackage{cite}
\usepackage{xcolor}
\usepackage{multirow}
\usepackage{array}
\usepackage{booktabs}
\usepackage{tabularx}
\usepackage[export]{adjustbox}
\usepackage{dblfloatfix}
\usepackage{xcolor}

\usepackage{natbib}
\bibpunct[, ]{(}{)}{;}{a}{}{,}
\renewcommand\bibfont{\fontsize{10}{12}\selectfont}

\theoremstyle{plain}

\theoremstyle{definition}

\theoremstyle{remark}

\usepackage{cleveref}

\usepackage[linesnumbered,ruled]{algorithm2e}

\SetCommentSty{mycommfont}

\usepackage[acronym]{glossaries}
\newacronym{fea}{FEA}{finite element analysis}
\newacronym{pmsm}{PMSM}{permanent magnet synchronous machine}
\newacronym{mtpa}{MTPA}{maximum torque per ampere}
\newacronym{cefea}{CE-FEA}{computationally efficient FEA}
\newacronym{ase}{ASE}{approximate solution evaluation}
\newacronym{ese}{ESE}{exact solution evaluation}
\makeglossaries

\newcommand{\bx}{\mbox{${\textbf x}$}}
\newcommand{\bux}{\mbox{${\textbf X}$}}
\newcommand{\buf}{\mbox{${\textbf F}$}}
\newcommand{\bug}{\mbox{${\textbf G}$}}

\newcommand{\fcand}{\buf^{\texttt{(cand)}}} 
\newcommand{\xcand}{\bux^{\texttt{(cand)}}} 
 
\newcommand{\xexploit}{\bux^{\texttt{(infill)}}} 
 
\newcommand{\fexploit}{\buf^{\texttt{(infill)}}} 

\newcommand{\ndoe}{N^{\texttt{DOE}}}

\begin{document}

\articletype{}

\title{Optimal Design of Electric Machine with Efficient Handling of Constraints and Surrogate Assistance}

\author{
\name{Bhuvan Khoshoo\textsuperscript{a}\thanks{Corresponding Author: Bhuvan Khoshoo. Email: khoshoob@msu.edu}, Julian Blank\textsuperscript{b}, Thang Q. Pham\textsuperscript{a}, Kalyanmoy Deb\textsuperscript{a}, \\and Shanelle N. Foster\textsuperscript{a}}
\affil{\textsuperscript{a}Department of Electrical and Computer Engineering, Michigan State University, East Lansing, USA; \textsuperscript{b}Department of Computer Science and Engineering, Michigan State University, East Lansing, USA}
}

\maketitle

\begin{abstract}
Electric machine design optimization is a computationally expensive multi-objective optimization problem. 
While the objectives require time-consuming finite element analysis, optimization constraints can often be based on mathematical expressions, such as geometric constraints. 
This article investigates this optimization problem of mixed computationally expensive nature by proposing an optimization method incorporated into a popularly-used evolutionary multi-objective optimization algorithm -- NSGA-II.
The proposed method exploits the inexpensiveness of geometric constraints to generate feasible designs by using a custom repair operator. 
The proposed method also addresses the time-consuming objective functions by incorporating surrogate models for predicting machine performance.
The article successfully establishes the superiority of the proposed method over the conventional optimization approach.
This study clearly demonstrates how a complex engineering design can be optimized for multiple objectives and constraints requiring heterogeneous evaluation times and optimal solutions can be analyzed to select a single preferred solution and importantly harnessed to reveal vital design features common to optimal solutions as design principles.

\end{abstract}


\section{Introduction}\label{sec:Intro}


Electric machines are essential components within a multitude of industries today and their range of application varies from refrigeration and industrial pumps to power generation and automobiles.
Consequently, design optimization of electric machines is a complex multi-objective optimization problem (MOOP) often involving combined analysis of electromagnetic, thermal, and structural performance.
Analysis of electric machines is a time-consuming process and therefore, a lot of effort has been focused on improving the optimization tools and efficiency in the past two decades.

From investigating pattern search and sequential unconstrained minimization techniques \citep{1973-ramarathnam-study, 1983-Bhim-sumt} to employing evolutionary algorithms (EAs) for optimization of electric machines \citep{1998-bianchi-direct-search}, electric machine community has continually adopted the improvements in optimization algorithms. 
As EAs perform better than classical direct search methods in finding global optimum their utilization in electric machine optimization has gained further popularity \citep{2002-Mirzaeian,Sudhoff2005,Zarko2005,Duan2009,Duan2013a,Zhang2013b}.
It is also worth mentioning that electric machines require \gls{fea} for performance evaluation with high accuracy.
Naturally, application of EAs in combination with FEA can also be found in literature \citep{2010-Pellegrino, 2010-pellegrino-multiobj}.
Although combination of EAs with FEA ensures finding optimal solutions with high quality, it also increases the overall computational cost of optimization.
In this regard, exploration of computationally inexpensive methods, such as surrogate models, to predict performance of electric machines has led to reduction in overall computational effort \citep{2005-Jolly-RSM,2009-ionel-surrogate,2012-sizov-ac,2018-taran-levelwise,2018-Song-DWKNN}.

A comprehensive literature review shows that while efforts have been made to improve the optimization tools and efficiency, constraints handling, specifically inexpensive constraints (such as geometric), in electrical machine optimization deserves more attention.
More commonly, the geometric feasibility of a candidate solution during optimization relies on random sampling.
For instance, in each optimization cycle (generation), geometrically infeasible solutions are discarded, and a random initialization may be repeated until the desired number of feasible solutions has been found~\citep{2015-stipetic-review}.
However, this random sampling may be inefficient when the number of geometric variables and constraints increases. \cite{Bhuvan} presented a preliminary study showing that the information from inexpensive constraints can be used to repair geometrically infeasible solutions and improve the Pareto-optimal front. However, this preliminary study did not address the computational expense of the objective functions vital for electric machine design optimization. Thus, this article extends the algorithm repairing infeasible designs by incorporating surrogate models to address the time-consuming objective functions. The main contributions of this work are as follows.

\begin{itemize}

\item Proposal of a repair operator which improves the quality of Pareto-optimal front by ensuring geometrically feasible solutions in each optimization cycle by exploiting the inexpensiveness of geometric constraints, while also respecting the manufacturing accuracy limitations.
\item Performance validation of the proposed repair operator in combination with surrogates to predict the computationally expensive objective functions and their impact on the convergence of the optimization algorithm.
\item Insights gained from Pareto-optimal electric machine designs and recommendations for selecting preferred solutions based on two different approaches: (1) a domain specific a posteriori multi-criteria decision-making (MCDM) method involving machine expertise, and (2) trade-off analysis of the Pareto-optimal set.

\end{itemize}

The rest of this article is structured as follows. Section~2 discusses related work and reviews optimization methods proposed to optimize electric machine design.
Section~3 discusses the formulation of the optimization problem used in this article.
Section~4 presents the proposed optimization method exploiting the computationally inexpensive constraints using a repair operator and addressing the computationally expensive objectives by incorporating surrogate models. 
The impact of the algorithm's components on the convergence of the algorithm, along with a detailed discussion about Pareto-optimal solutions and selection of preferred electric machine designs, is discussed in Section~5. Finally, conclusions are presented in Section~6.

\section{Related Work}

Advancements in optimization algorithms and objective function evaluation tools have facilitated the design optimization of electric machines.
\cite{1973-ramarathnam-study} presented an early case study involving an induction machine's design by solving a single objective optimization problem.
The authors compared the performance of a direct, indirect, and random search method in conjunction with the sequential unconstrained minimization technique.
Results showed that a direct search method performs better for complicated multi-variable functions commonly occurring in electric machines. However, optimization methods considered in the study suffered from getting stuck in local optima and required several restarts to reach the global optimum.

Metaheuristics, particularly Genetic Algorithms (GAs), are widely used and known for their global search behavior. 
\cite{1998-bianchi-direct-search} used a GA to optimize the design of a surface-mounted permanent magnet (SPM) machine. 
Results from two independent single objective optimization problems indicated that an evolutionary method outperforms the direct search method when comparing the convergence to the global optimum.

Since the design of an electric machine typically includes comparing the performance of multiple metrics, multi-objective optimization using EAs is predominantly employed nowadays.
For instance, \cite{2010-pellegrino-multiobj} employed an EA combined with \gls{fea} to solve a three-objective optimization problem.
The authors compared two partial optimization strategies with a comprehensive three-objective optimization method.
Results showed that domain knowledge could be utilized to modify the optimization problem creatively to reduce the computation time without significantly affecting the quality of results.


Several other strategies have been proposed for the reduction of optimization run-time. 
For example, \cite{2015-pellegrino-automatic} proposed a local refinement strategy to improve a Pareto-optimal design further after the optimization terminated. After selecting a design in the region of interest, a local optimization method was employed in the design's vicinity.
Results showed that an a posteriori local search, even with fewer function evaluations, produced similar results to those by an approach solely relying on global optimization.
Similarly, \cite{2016-degano-global} split the optimization procedure into two phases, where authors optimized torque density and losses in the first stage and the quality of torque profile in the second stage.
Although average torque and ripple are conflicting objectives, the optimal solutions after the second stage showed improved torque ripple without compromising the average torque.

Another research direction to address computationally expensive functions during optimization is the usage of surrogate models.
For example, \cite{2005-Jolly-RSM} used second-order response surface models (RSM) to predict d-axis and q-axis inductances and magnet flux linkage of an IPM machine to optimize the magnet shape and placement in the machine's rotor.
Similarly, \cite{2018-taran-levelwise} presented a two-level surrogate-assisted optimization approach using DE to find optimal designs for Axial Flux PM (AFPM) machines by minimizing active material mass and total losses at rated operation. Results showed that the surrogate-assisted algorithm outperforms the conventional multi-objective DE in terms of computation time.
It is worth mentioning that usage of surrogates introduces a trade-off between computation time and solution accuracy.

It is clear from the review of past studies that constraint handling, especially when the constraints are inexpensive, in an electric machine design optimization problem requires more attention.
This article addresses this gap in research by first introducing an electric machine design problem of mixed computationally expensive nature and then proposes an optimization method that exploits the inexpensiveness of constraints.

\section{Electric Machine Design And Optimization Problem Formulation}

\begin{figure}
\centering
\hspace*{\fill}%
\subfigure[2d model of machine.\label{fig:full model}]
{\includegraphics[width=0.45\textwidth, height = 4.5cm, keepaspectratio]{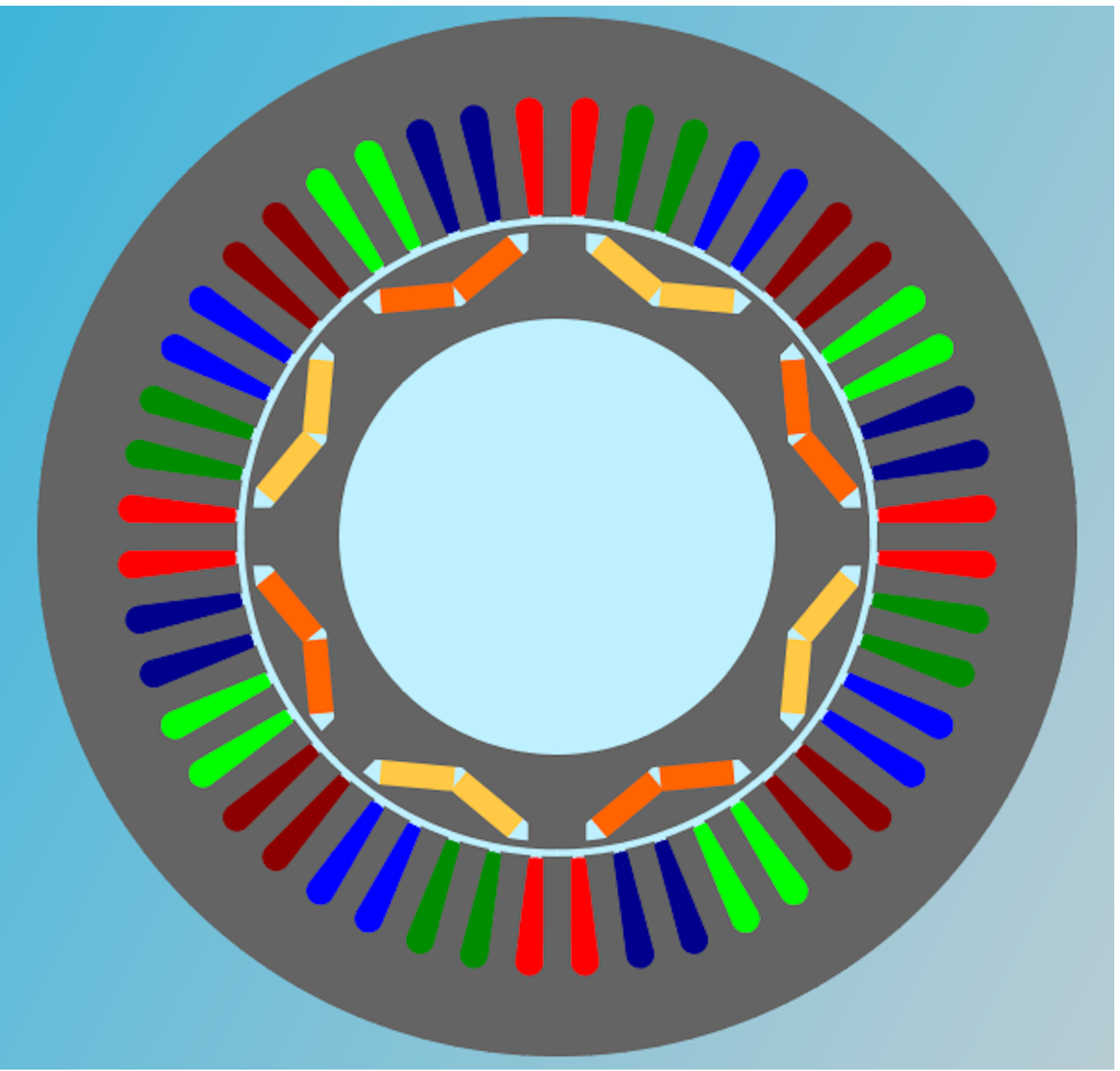}}
\hfill
\subfigure[Reduced model used in FEA.\label{fig:reduced model}]
{\includegraphics[width=0.45\textwidth, height = 4.5cm, keepaspectratio]{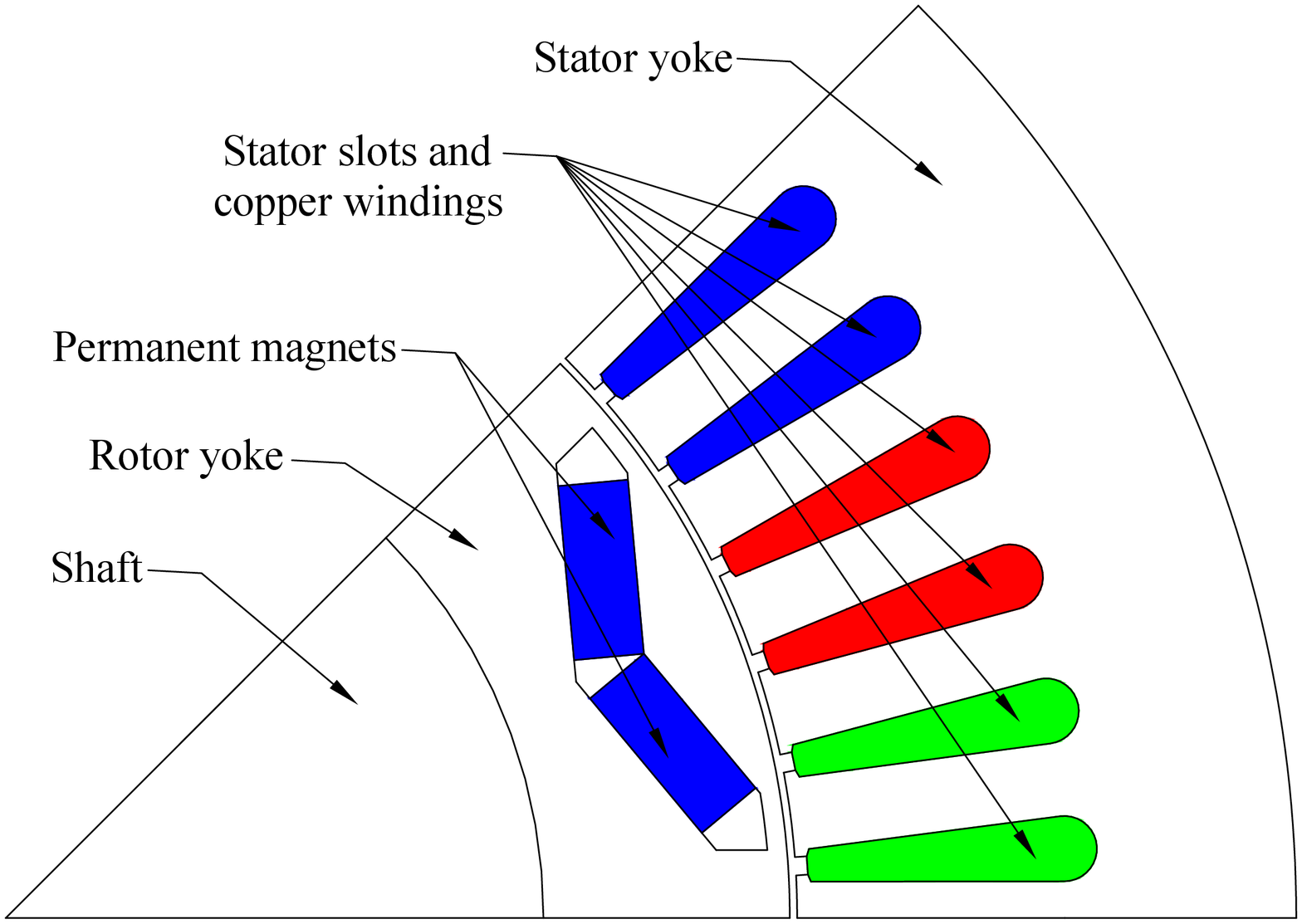}}
\hspace*{\fill}%
\caption{IPM machine used for optimization} 
\label{fig:electric-2d}
\end{figure}

\begin{figure}
\hspace*{\fill}%
\begin{minipage}{0.43\textwidth}
    \centering
    \includegraphics[width=0.99\textwidth, keepaspectratio]{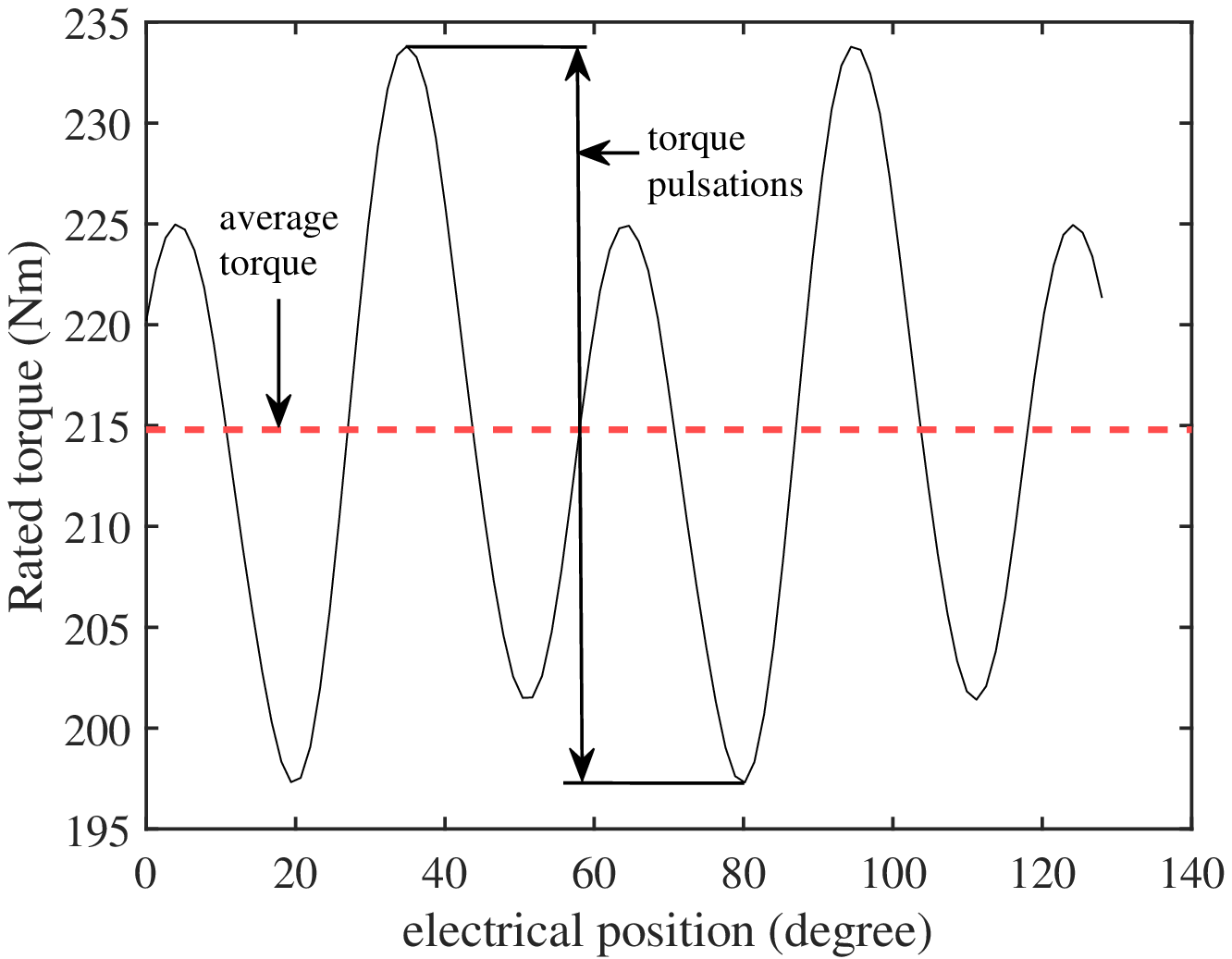}
    \caption{ \label{fig:torque-ref} Torque profile of reference mahine at rated operation.}
\end{minipage}
\hspace{4mm}
\begin{minipage}{0.45\textwidth}
\centering
    \includegraphics[width=0.93\textwidth, keepaspectratio]{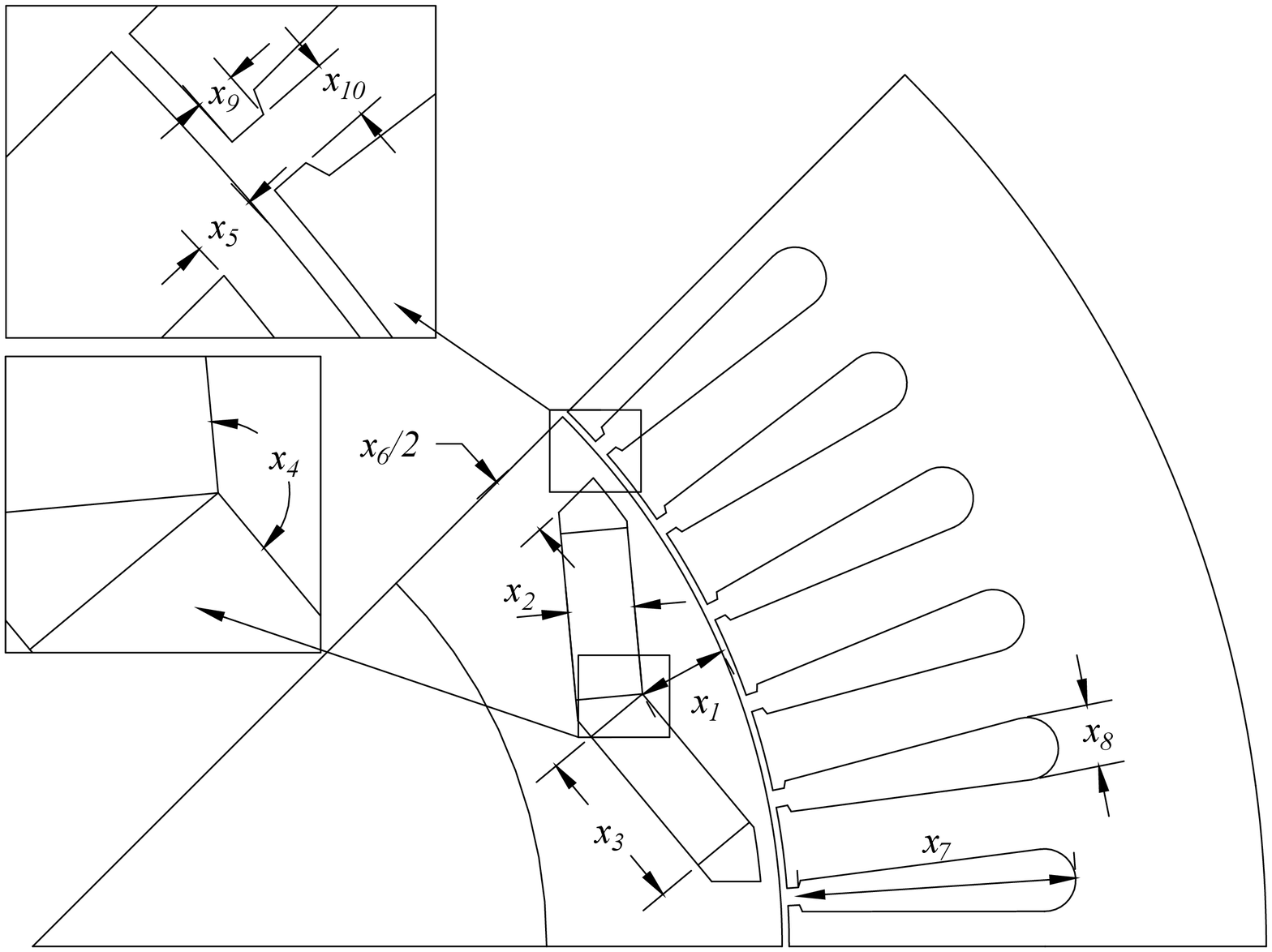}
    \caption{ \label{fig:geom-var} Geometric variables used for optimization.}
\end{minipage}
\hspace*{\fill}%
\end{figure}

In addition to selection of objective functions, design variables, variable ranges, and constraints like every other MOOP, an electric machine design optimization problem also requires selection of a machine template which is primarily application dependant.
In this article, a 3-phase, 48-slot/8-pole IPM machine with a single layer of V-shaped magnet, used in 2010 Toyota Prius, is chosen for analysis.
2D model of the machine is shown in Figure~\ref{fig:full model} \citep{Altair-fluxmotor}.
In this article, two of the most common machine performance measures, average torque and torque pulsations are chosen as the objective functions which are calculated using FEA. To reduce the simulation run time, periodicity in 2D model is taken advantage of and only $1/8^{th}$ of the model is simulated as shown in Figure~\ref{fig:reduced model}.
Both objective functions are conflicting and optimization's goal is to maximize average torque while minimizing torque pulsations, where the definition of pulsations is highlighted in Figure~\ref{fig:torque-ref}.

After the selection of objective functions, a sensitivity analysis study has provided the ten most significant geometric variables, as shown in Figure~\ref{fig:geom-var}. 
Variable ranges are defined based on the machine designer's experience with a $\pm$20\% variation from the reference design.
Additionally, manufacturing accuracy limitations are applied to all variables by limiting them to have only two decimal places.
Details of lower ($x^{(L)}$) and upper ($x^{(U)}$) bounds along with reference ($x^{(ref)}$) values of the ten variables are given in Table~\ref{tbl:electric-variables}.
Ten geometric constraints ensure the geometric feasibility of candidate designs.
All optimization studies are performed for the rated operating point, with the rotational speed of the rotor and the excitation angle kept constant to those of the reference design.
Details on the formulation of geometric constraints and the selection of an operating point for optimization are provided in the supplementary document.

\begin{table}
\tbl{Values of geometric variables used for optimization.}
{\begin{tabular}{lllccc} \toprule
 $x_i$ & Variable & Unit & $x^{(\text{ref})}$ &  $x^{(L)}$ &	$x^{(U)}$ \\ \midrule
$x_1$ & Height of rotor pole cap  & mm &	9.56	& 7.65	& 11.47\\
$x_2$ & Magnet thickness & mm &	7.16 & 	5.73 &	8.59\\
$x_3$ & Magnet width & mm &	17.88 &	14.30 &	21.46\\
$x_4$ & Angle between magnets & degree &	145.35 &	116.28	& 174.42\\
$x_5$ & Bridge height & mm &	1.99 &	1.59 &	2.39\\
$x_6$ & Q-axis width & mm &	13.9 &	11.12 &	16.68\\
$x_7$ & Slot height & mm &	30.9 &	24.72 &	37.08\\
$x_8$ & Slot width & mm	& 6.69 &	5.35 &	8.03\\
$x_9$ & Height of slot opening & mm	 & 1.22 &	0.98 &	1.46 \\
$x_{10}$ & Width of slot opening & mm &	1.88 &	1.50 &	2.26\\ \bottomrule
\end{tabular}}
\label{tbl:electric-variables}
\end{table}

Based on the above discussion of electric machine design, a bi-objective optimization problem with ten variables and ten geometric constraints is formulated in this work.
Ultimately, the MOOP is defined as
\begin{equation}
\begin{aligned}
\text{Maximize}~~  & \quad \texttt{Average\ Torque}(\bx), \\
\text{Minimize}~~\,  & \quad \texttt{Torque\ Pulsation}(\bx), \\
\text{subject~to}~~ & \quad g_j(\bx) \leq 0, & \forall j\in 1,\ldots,10,~~\\
& \quad x_i^{(L)} \leq x_i \leq x_i^{(U)}, & \forall i\in 1,\ldots,10,~\\
\text{where}~~  & \quad \bx \in \mathbb{R}^N,
\end{aligned}
\label{eqn:electric-problem}
\end{equation}
where $\bx$ represent the design variables to optimize, $g_j(\bx)$ are the geometric constraints, and the lower and upper bound of the variables are denoted by $x_i^{(L)}$ and $x_i^{(U)}$ respectively. As explained previously, due to manufacturing accuracy limitations, all variables are restricted to have only two decimal places.
Additionally, while the geometric constraints are inexpensive to evaluate, objective functions require time consuming FEA.

\section{Proposed Multi-Objective Optimization Algorithm}

Based on the discussion presented in the previous section, it can be concluded that the formulated electric machine optimization problem is of mixed computationally expensive nature with two expensive objective functions and ten inexpensive geometric constraints.
In a preliminary study, \cite{Bhuvan} showed that the computational inexpensiveness of constraint evaluations could be exploited to convert an infeasible solution to a feasible one through a repair operator.
Additionally, the design optimization of electric machines is an expensive problem to solve, and some effort must be made to reduce the computational cost.
Therefore, in addition to the repair operator, the proposed optimization method incorporates surrogates for predicting expensive objectives.
In this study, the evolutionary multi-objective optimization (EMO) algorithm NSGA-II~\citep{Deb-NSGA2} is used as the base optimization algorithm.
Implementation of repair operator and surrogates in optimization algorithm is explained below.

\subsection{Implementation of Repair Operator}

Implementation of repair operator focuses on two goals: (1) converting an infeasible solution to a feasible one and (2) satisfying the manufacturing accuracy limitations.
While the first goal is achieved via an embedded optimization procedure, the second goal is accomplished by rounding each variable up or down, using an approach inspired by Hooke-Jeeves pattern moves \citep{hooke-jeeves}.
A detailed explanation of the two phases is included in the supplementary document.

\subsection{Surrogate Incorporation in Optimization Cycle}


Commonly, surrogates -- approximation or interpolation models -- are utilized during optimization to improve the convergence behavior. 
First, one shall distinguish between two different types of evaluations: \glspl{ese} that require to run the computationally expensive evaluation for computing two objectives \texttt{Average\ Torque}(\bx) and \texttt{Torque Pulsation}(\bx); and \glspl{ase} which is a computationally inexpensive approximation by the surrogates. 
Where the overall optimization run is limited by $\texttt{ESE}^{\max}$ function evaluation, function calls of \glspl{ase} are only considered as algorithmic overhead. 
In order to improve the convergence of NSGA-II, the surrogates provide \glspl{ase} and let the algorithm look several iterations into the future without any evaluation of \glspl{ese}. 
The surrogate models are used to create a set of infill solutions as follows: First, NSGA-II is run for $k$ more iterations (starting from the best solutions found so far), returning the solution set $\xcand$.
The number of solutions in $\xcand$ corresponds to the population size of the algorithm fixed to $100$ solutions in this study.
After eliminating duplicates in $\xcand$, the number of solutions $N$ desired to run using \glspl{ese} needs to be selected. The selection first obtains $N$ clusters by running the k-means algorithm and then uses a roulette wheel selection based on the predicted crowding distances. Note that this will introduce some selection bias towards the boundary points as they have been depicted with an infinite crowding distance.
Altogether, this results in $N$ solutions to be evaluated using \glspl{ese} in the current optimization cycle.

\begin{figure}[hbt]
    \centering
    \includegraphics[width=0.5\textwidth, keepaspectratio, trim=0.3cm 7.7cm 21.6cm 0,clip]{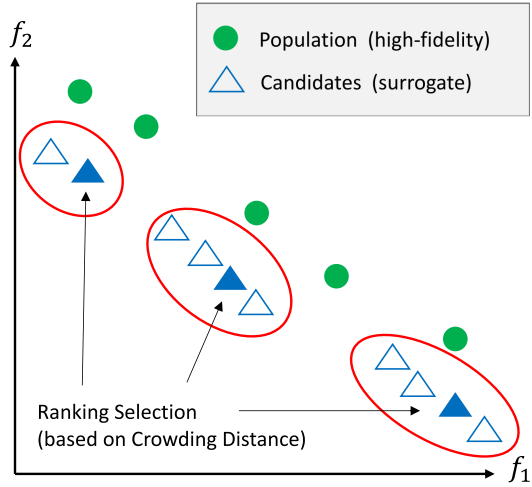}
    \caption{Ranking selection of solutions obtained by optimizing the surrogate-based optimization problem.}
    \label{fig:surrogate}
\end{figure}

Since the electric machine design is formulated with two objectives, two different models are built. Separately fitting a model for each objective corresponds to the \emph{M1} method proposed in the surrogate usage taxonomy~\citep{2019-deb-taxonomy}.
For each objective, the best model type is found by iterating over different model realizations of  RBF~\citep{1971-hardy-rbf} and Kriging~\citep{1951-krige-mine} varying normalization, regression, and kernel type. Finally, the best model type is chosen based on the validation set's performance.

\subsection{NSGA-II-WR-SA}

Algorithm~\ref{alg:pseudo-code} shows how the repair operator and surrogate models are incorporated into the optimization cycle. The algorithm's parameters are the expensive objective functions \buf(\bux) and the inexpensive constraint functions \bug(\bux); the maximum number of exact solution evaluations $\texttt{ESE}^{\max}$ serves as an overall termination criterion; the number of the initial design of experiments $\ndoe$ describes how many designs are evaluated before optimization starts; the number of solutions $N$ evaluated in each optimization cycle; and the number of surrogate optimization generations $k$, or in other words, how many generations the surrogates are used to look into the future.

\begin{algorithm}[t]
\small
\SetNoFillComment
\SetKwInOut{Input}{Input}
\Input{Expensive Objective Functions \buf(\bux), 
Inexpensive Constraint Function \bug(\bux), 
Maximum Number of Exact Solution Evaluations $\texttt{ESE}^{\max}$, 
Number of Initial Design of Experiments $\ndoe$,
Number of \glspl{ese} in Each Optimization Cycle $N$,  
Number of Surrogate Optimization Generations $k$.
}

\vspace{1mm}

\tcc{initialize feas. solutions using the inexpensive function $G$}

$\bux \gets \texttt{constrained\_sampling}(\ndoe, \bug)$
\label{alg:init}

$\buf \gets \buf(\bux)$
\label{alg:init_eval}

\vspace{1mm}

\While{$|\bux| < \texttt{ESE}^{\;\max}$}{
\label{alg:while}

    \vspace{1mm}
    \tcc{exploitation using the surrogate}
    $\hat{\buf} \gets \texttt{fit\_surrogate}(\bux, \buf)$
    \label{alg:surrogate}
    
    $\left(\xcand, \fcand\right) \gets \texttt{optimize}(\texttt{'NSGA-II-WR'}, \hat{\buf}, \bug, \bux, \buf, k)$
    \label{alg:surrogate_optimize}
    
    $\left(\xcand, \fcand\right) \gets \texttt{eliminate\_duplicates}(\bux, \xcand, \fcand)$
    \label{alg:dupl_elim}
    
    $C \gets \texttt{cluster}(\texttt{'k\_means'}, N^{(\texttt{exploit})}, \fcand)$
    \label{alg:cluster}
    
    $\xexploit  \gets \texttt{ranking\_selection}(\xcand, C, \, \texttt{crowding}(\fcand))$
    \label{alg:selection}
    
    \vspace{1mm}
    \tcc{evaluate and merge to the archive}
    $\fexploit  \gets \buf(\xexploit );$
    \label{alg:eval}
    
    $\bux \gets \bux  \cup \xexploit $
    
    $\buf \gets \buf \cup \fexploit $
    \label{alg:merge_f}

\vspace{1mm}
}

\caption{NSGA-II-WR-SA: A customized version of NSGA-II with a repair of infeasible solutions (WR) and surrogate assistance (SA).}
\label{alg:pseudo-code}
\end{algorithm}

First, the algorithm starts by sampling $\ndoe$ solutions in the feasible space using the constrained sampling strategy
(Line~\ref{alg:init})~\citep{2021-blankjul-cheap-constr} and evaluates the solution set  (Line~\ref{alg:init_eval}).
Then, while the overall evaluation budget $\texttt{ESE}^{\max}$ has not been used yet, surrogates $\hat{\buf}$ are built for the objectives (Line~\ref{alg:surrogate}).
By applying NSGA-II for $k$ surrogate optimization generations starting from \bux, using the surrogate models $\hat{\buf}(\bux)$ and the inexpensive objective functions \bug(\bux), a candidate set of solutions $\xcand$ and $\fcand$ is retrieved (Line~\ref{alg:surrogate_optimize}). 
Depending on the surrogate problem, some solutions in $\xcand$ can be identical to the ones already evaluated in \bux; thus, duplicate elimination is necessary to ensure these solutions are filtered out (Line~\ref{alg:dupl_elim}).
Since the size of $\xcand$ exceeds $N$, a subset solution based on the predicted crowding distances takes place~(Line~\ref{alg:cluster} and~\ref{alg:selection}). 
Finally, the resulting solution set $\xexploit$ of size $N$ is evaluated using \glspl{ese} and is appended to the archive of solutions.

\section{Results and Discussion}

In this section, the performance of the proposed optimization method is investigated and following key questions are answered.

\begin{itemize}
    \item How does the repair operator help the optimization cycle and what is its impact on the Pareto-optimal front?
    \item Does the usage of surrogates improve the convergence behavior of the proposed optimization method?
    \item What can be learned from the Pareto-optimal solutions, each representing an electric machine design?
\end{itemize}

\begin{table}
\tbl{Optimization Setup and Results for all five runs combined for NSGA-II and NSGA-II-WR. HV is calculated after normalization of objective functions.}
{\begin{tabular}{lcccccc} \toprule
Algorithm && Description & Evals & Feasible & Non-dominated & HV \\ \midrule
NSGA-II && Conventional & 7,500 & 5,446 & 27 & 0.7206\\
NSGA-II-WR && With Repair & 7,500 & $\mathbf{7,500}$ & $\mathbf{59}$ & $\mathbf{0.7382}$\\
\bottomrule
\end{tabular}}
\label{tbl:Optimization-setup}
\end{table}

\subsection{Impact of Repair Operator}

\begin{figure}
\centering
\subfigure[NSGA-II (7,500 evals).\label{fig:NSGA-II}]{
\includegraphics[width=0.48\textwidth, trim=0 0 20 10, clip]{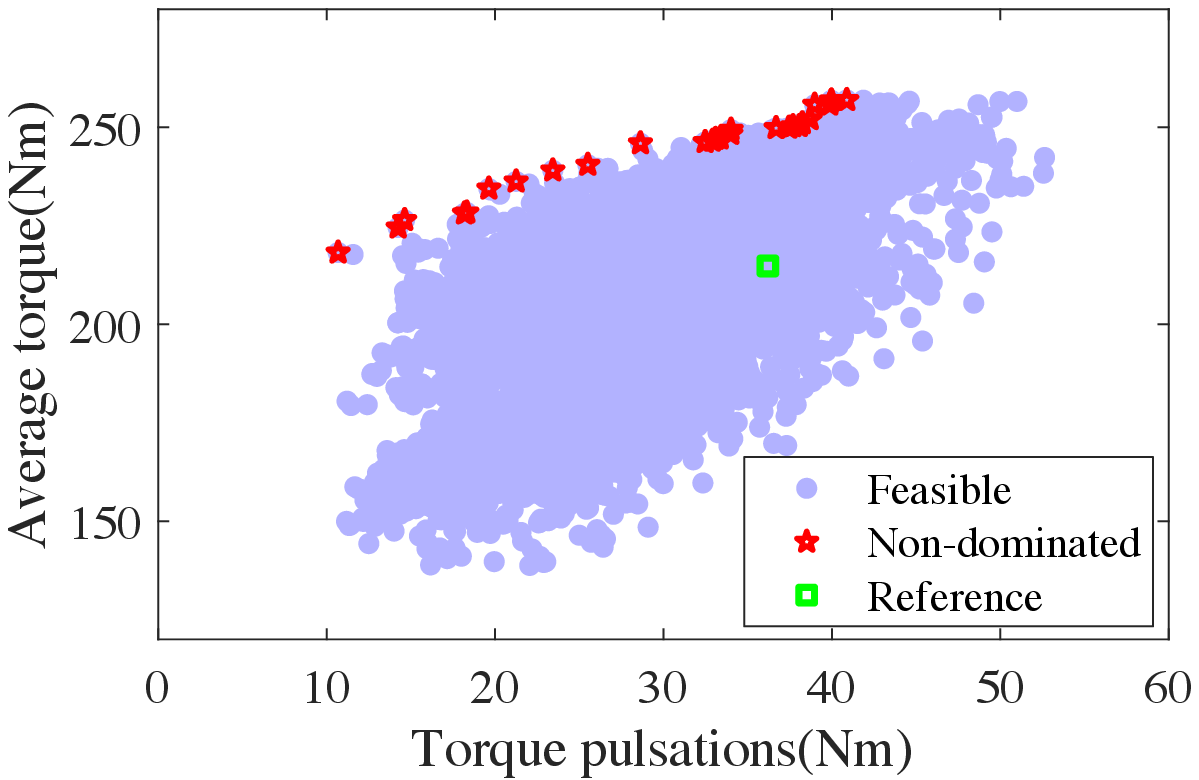}}
\subfigure[NSGA-II-WR (7,500 evals).\label{fig:NSGA-II-WR}]{
\includegraphics[width=0.48\textwidth, trim=0 0 20 10, clip]{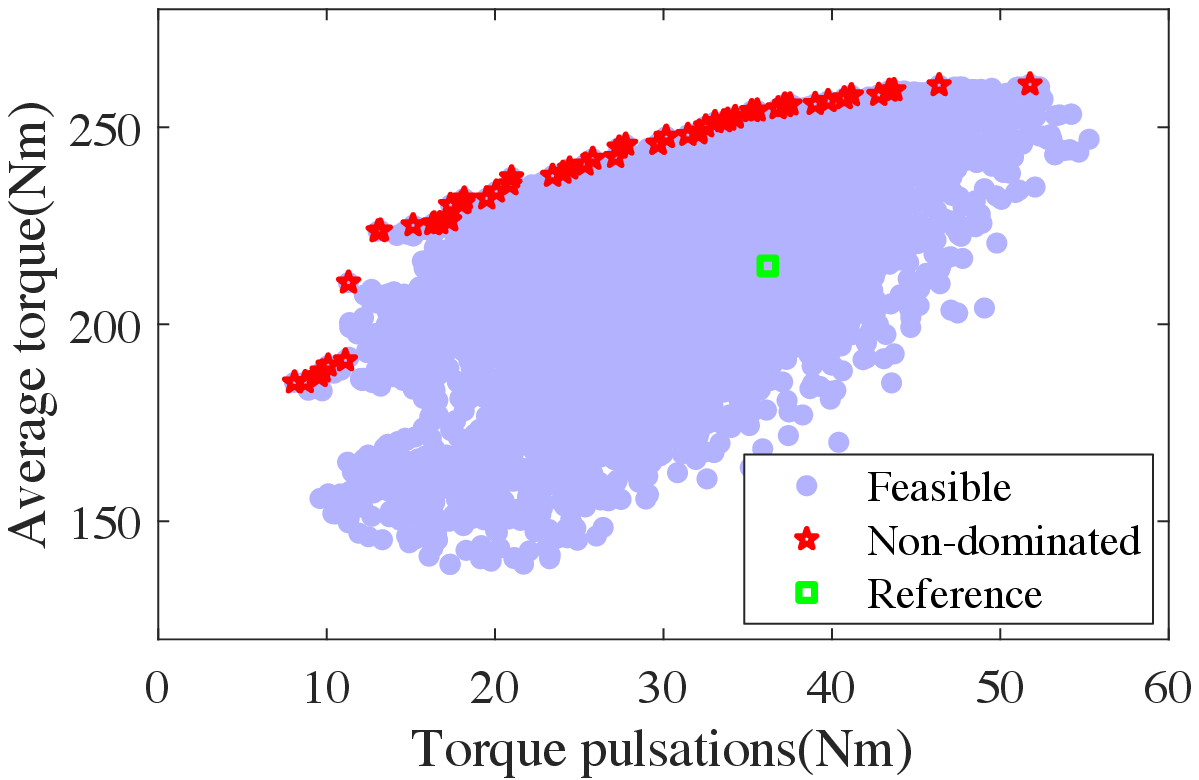}}
\caption{Objective space illustrating dominated and non-dominated (Pareto) solutions obtained with all runs combined of NSGA-II and NSGA-II-WR algorithms in \subref{fig:NSGA-II} and \subref{fig:NSGA-II-WR}, respectively. All feasible solutions obtained since the start of a specific run are shown in blue color.}
\label{fig:results-optim}
\end{figure}

It is helpful first to analyze the constraints formulated in this article to understand the impact of the repair operator. A preliminary study with 10,000 randomly sampled solutions shows that only $30.3\%$ of samples are feasible without violating any of the ten geometric constraints. Further details about the analysis of constraints are included in the supplementary document.
This article investigates the impact of the repair operator by comparing two optimization methods: (1) the conventional NSGA-II and (2) NSGA-II combined with the repair operator called NSGA-II-WR in the rest of the article. Both methods use simulated binary crossover (SBX) operator with a probability of 0.9 and polynomial mutation along with binary tournament selection. The distribution indices used in this study are set to $\eta_c = 15$ and $\eta_m = 20$ for crossover and mutation operators respectively.
For each method, five optimization runs with different seeds are completed. However, the seeds are kept the same for the two methods for a fair comparison. Each optimization run consists of 1500 total evaluations ($\texttt{ESE}^{\max}=1500$), with a population size of 100 and 20 offsprings. The two optimization methods are compared based on the combined results of the five runs, and the overall setup and details are shown in Table~\ref{tbl:Optimization-setup} and Figure~\ref{fig:results-optim}. It is clear that the use of the repair operator yields more non-dominated solutions and also results in a Pareto-optimal front with larger hypervolume (HV) \citep{Hypervolume-Zitzler} than the conventional method.
For the calculation of HV, the worst and the best points are found from the combined set of the two Pareto-optimal fronts.
After that, the objective functions are normalized to obtain the normalized HV.
Additionally, comparisons of individual runs from the two methods show that the Pareto-optimal fronts obtained with NSGA-II are discontinuous and mostly dominated by those obtained with NSGA-II-WR.
A comparison of the runs with median HV and the best HV is included in the supplementary document.
Ultimately, the proposed repair operator is well suited for the design optimization of electric machines.

\subsection{Convergence Analysis With and Without Surrogates}

\begin{figure}
\centering
\subfigure[\label{fig:NDS-comparison}]{
\includegraphics[width=0.48\textwidth, trim=0 0 20 10, clip]{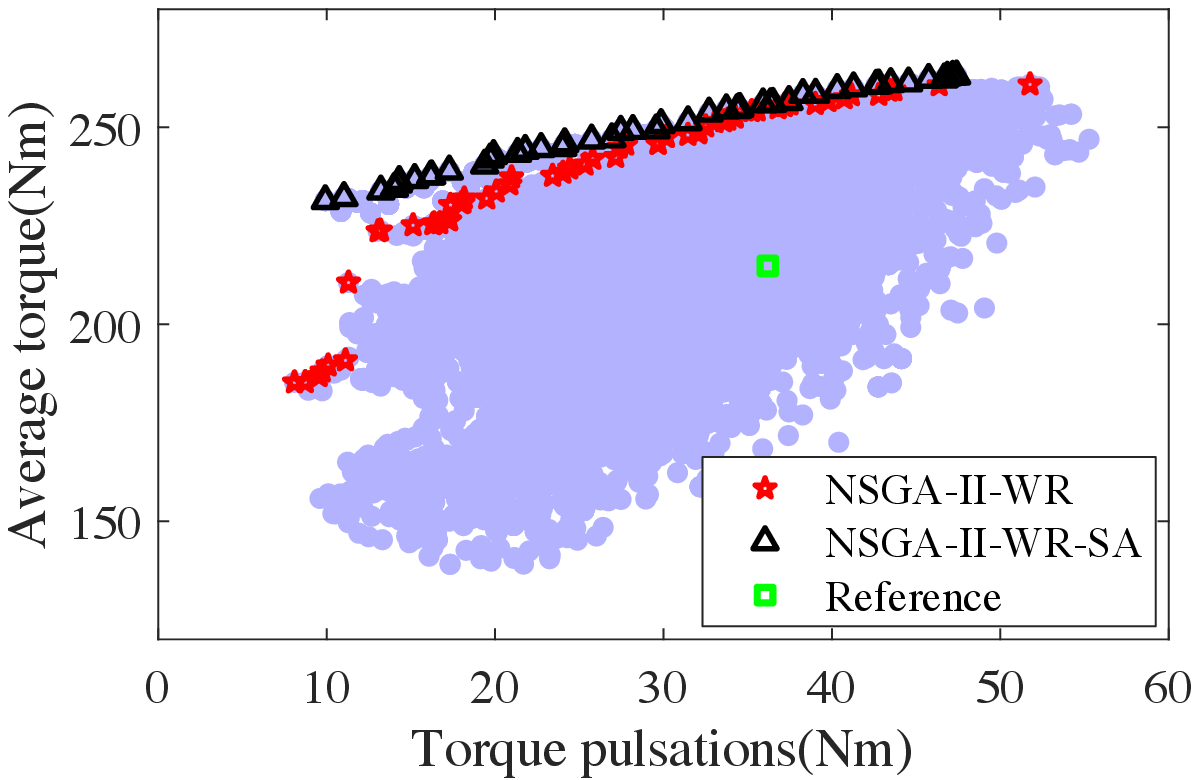}}
\subfigure[\label{fig:Variable-pcp}]{
\includegraphics[width=0.48\textwidth]{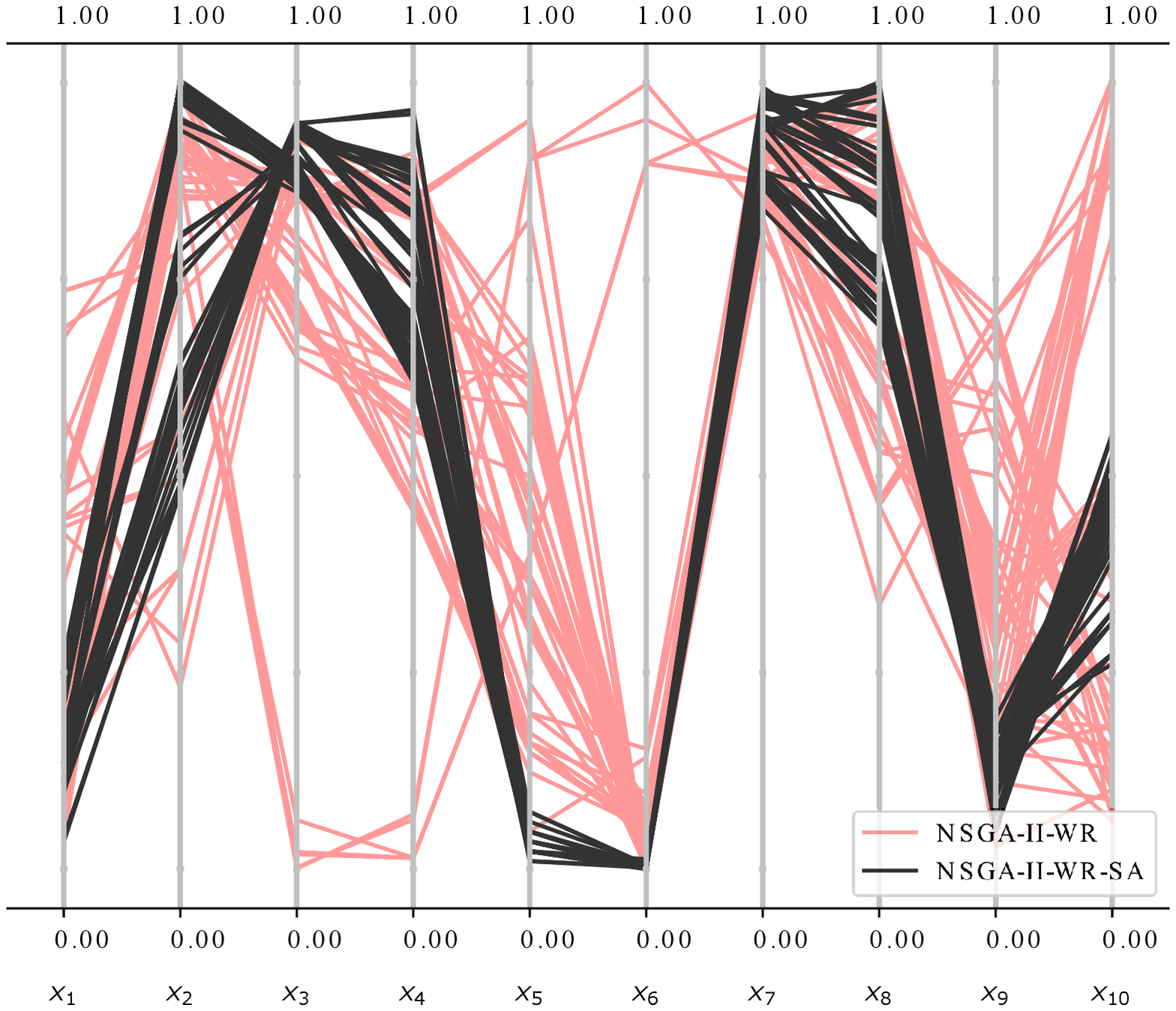}}
\caption{Comparison of objective space with Pareto-optimal fronts and normalized design space of Pareto optimal sets in \subref{fig:NDS-comparison} and \subref{fig:Variable-pcp} respectively. Pareto-optimal sets are obtained by combining all runs completed with the optimization methods, NSGA-II-WR and NSGA-II-WR-SA, respectively.}
\label{fig:NDS-final}
\end{figure}

Although surrogate-assisted optimization is known to find the Pareto-optimal front quicker than other methods, it is also sensitive to (model and optimization-related) hyperparameters.
In this article, the following three hyperparameters are varied to analyze the performance of the proposed optimization method with surrogates.
\begin{itemize}
    \item $N$: Number of \glspl{ese} in each iteration
    \item $k$: Number of surrogate optimization generations for exploitation
    \item $N^{\texttt{DOE}}$: Number of initial design of experiments
\end{itemize}
The complete setup and results of this parametric study and some important observations are included in the supplementary document for reference.
Based on this study, $N$ = 10, $k$ = 35, and $N^{\texttt{DOE}}$ = 60, is identified as the best parameter setting out of the analyzed configurations.
For the remainder of this article, the corresponding surrogate-assisted optimization configuration is referred to as NSGA-II-WR-SA, and its results are compared with those obtained with NSGA-II-WR.
A comparison of the two Pareto-optimal fronts clearly shows that NSGA-II-WR-SA outperforms NSGA-II-WR, as shown in Figure~\ref{fig:NDS-comparison}.
It should be noted that the compared Pareto-optimal fronts are obtained by combining all five runs for both algorithms, NSGA-II-WR and NSGA-II-WR-SA, respectively.
To understand the convergence of each optimization method, the design space of the two Pareto-optimal sets are visualized by a parallel coordinates plot (PCP) (see~Figure~\ref{fig:Variable-pcp}).
Each vertical axis in the PCP represents the normalized variable $x_i$ with its lower and upper bounds as 0 and 1, respectively, and each horizontal line represents a solution.
The design space of the two Pareto-optimal sets shows that most of the variables have converged to an optimal value with NSGA-II-WR-SA, whereas, with NSGA-II-WR, some of the variables still have significant variations with further scope for improvement.
These observations validate the effectiveness of the incorporation of surrogates by demonstrating the improvement in the convergence.
Moreover, it should be noted that while NSGA-II-WR has used 7,500 evaluations in this experiment, NSGA-II-WR-SA has converged to a better set of solutions with a solution evaluation budget of only 1,000.
As explained in the previous section, surrogates are utilized to look $k$ generations (here $k=35$) into the future to generate $N$ number of infill solutions in each optimization cycle, which leads to better convergence.
A comparison of individual runs included in the supplementary document further demonstrates the superiority of NSGA-II-WR-SA over NSGA-II-WR.
Additionally, a discussion on the performance of surrogates and exploration of search space included in the supplementary document shows that the convergence with surrogates depends on the complexity of the objective functions under consideration.

\subsection{Analysis of Pareto-Optimal Solutions}

The design space of the Pareto-optimal set obtained with NSGA-II-WR-SA is investigated to gain insights into the electric machine design (see~Figure~\ref{fig:Variable-pcp}). In general, while machine flux linkages affect the average torque, magnet pole arc and material saturation control the torque pulsations. Nevertheless, some critical observations are listed below.

\begin{itemize}
    \item Most of the Pareto-optimal solutions have larger values of magnet width ($x_3$), which results in more magnet flux linkage and average torque. Larger magnet width also results in smaller q-axis width ($x_6$).
    \item Most solutions also have larger values of slot height ($x_7$) and slot width ($x_8$) and, therefore, slot cross-section. A larger slot cross-section area results in more winding space, which translates to higher allowable excitation current and an increase in average torque.
    \item Reduction in bridge height ($x_5$) directly increases the air-gap flux density, which increases average torque.
    \item Magnet pole arc is directly proportional to magnet width ($x_3$) and angle between the magnets ($x_4$), and an increase in magnet pole arc seems to reduce pulsations.
    \item Material saturation is a nonlinear behavior observed in magnetic materials, such as electrical steel, introducing saturation harmonics in magnetic flux density. While a larger slot cross-section increases average torque through more excitation current, it also increases magnetic material saturation, leading to more torque pulsations.
    \item Lastly, the height and width of slot opening, $x_9$ and $x_{10}$, which are responsible for slot harmonics, have converged to the lower end of the variable range.
\end{itemize}

\subsection{Selection of Preferred Solutions}

The selection of an electric machine design is primarily application-dependent.
One approach could be to use a scalarized function yielding a single optimal solution \citep{Scalarized-Islam}.
However, proper scalarization of objectives is a difficult task.
Scalarization also does not offer the possibility of analyzing trade-offs observed for multiple objectives.
Moreover, optimizing all objectives likely produces a Pareto-optimal front which is harder to interpret and gain insights into the electric machine design.
This article uses two approaches to select the preferred solutions: (1) a domain-specific a posteriori MCDM method that involves machine expertise and (2) a trade-off analysis of the Pareto-optimal set to identify and choose the solutions with the highest trade-off.
Pareto-optimal solutions obtained from combined runs of NSGA-II-WR and NSGA-II-WR-SA optimization methods are used in both approaches.

\subsubsection{Domain Specific A Posteriori MCDM Method}

For a domain-specific a posteriori MCDM method, three performance measures, in addition to the two objective functions defined in~\eqref{eqn:electric-problem}, are used to select preferred solutions from the Pareto-optimal set.

\begin{itemize}
    \item Total harmonic distortion of noload back emf (\emph{THDV})
    \item Peak of fundamental of back emf (\emph{F-BEMF})
    \item Magnet utilization factor (\emph{MUF})
\end{itemize}

\begin{figure}
\centering
\subfigure[\label{fig:Pareto-THDV}]{
\includegraphics[width=0.48\textwidth]{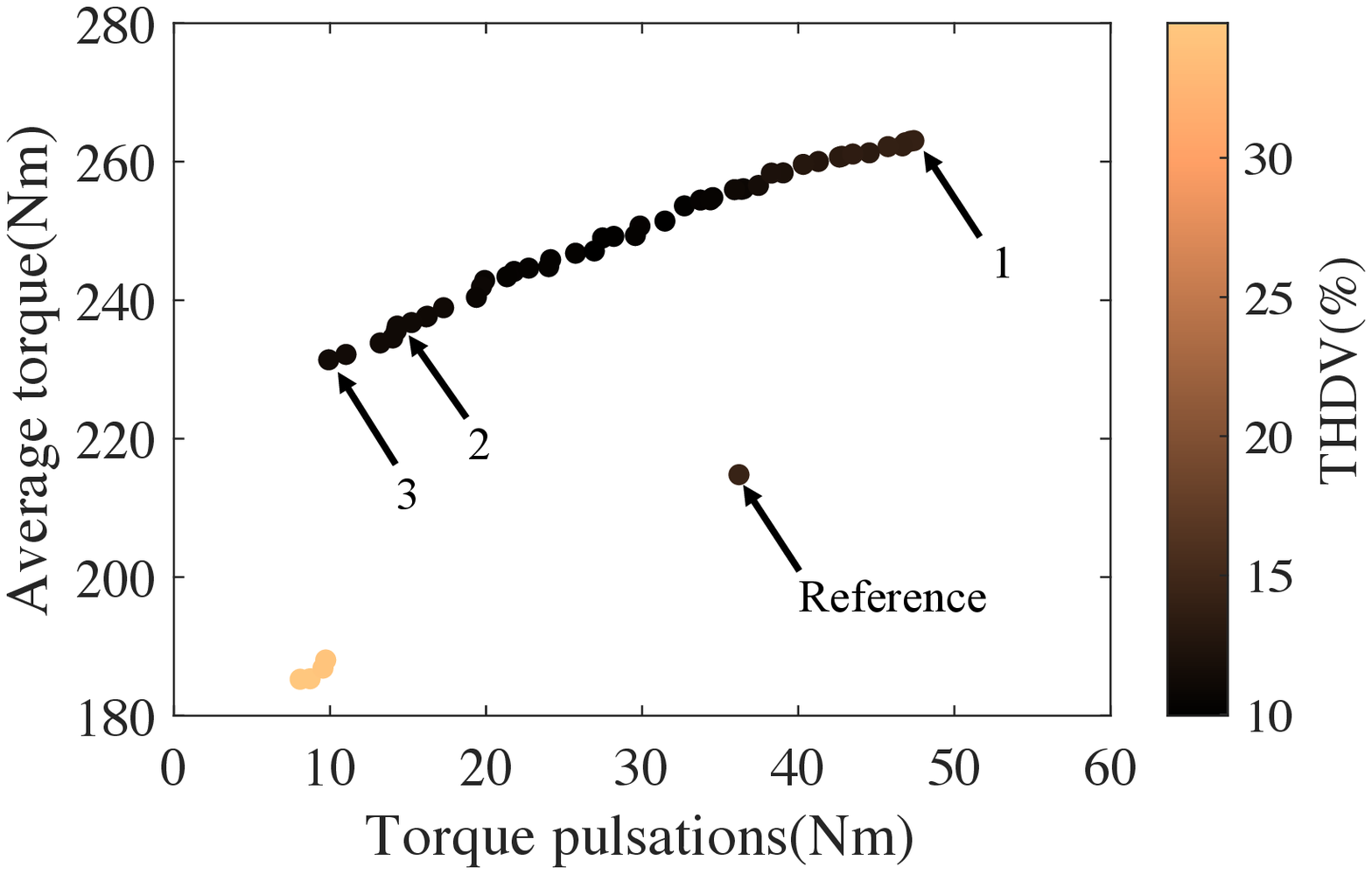}}
\hfill
\subfigure[\label{fig:Pareto-Trade-off}]{
\includegraphics[width=0.45\textwidth]{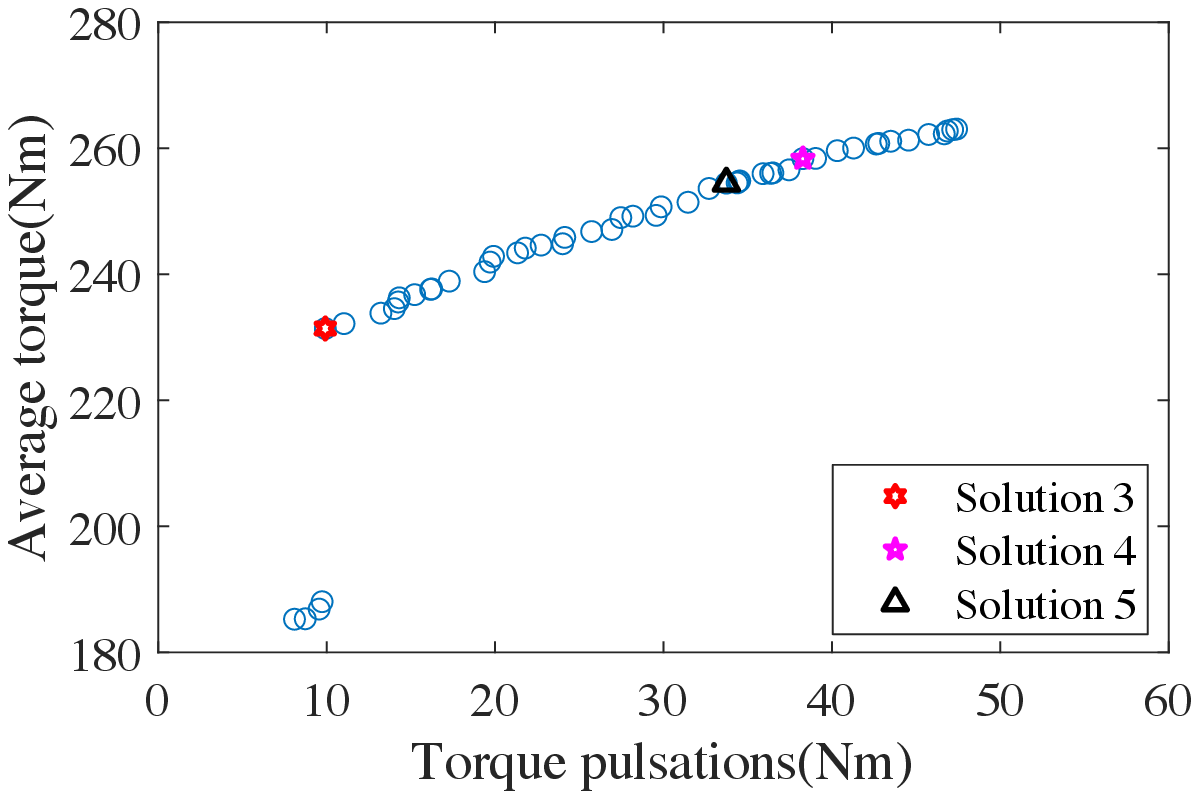}}
\caption{Objective space highlighting the selected solutions using the a posteriori MCDM method and trade-off analysis in \subref{fig:Pareto-THDV} and \subref{fig:Pareto-Trade-off}, respectively.} 
\label{fig:MCDM-selection}
\end{figure}

Since \emph{THDV} is directly proportional to noise, vibration, and harshness (NVH) during the operation of an electric machine, a solution with low \emph{THDV} is desirable.
Conversely, \emph{F-BEMF} instead introduces a trade-off as a high \emph{F-BEMF} increases average torque, but it also leads to a reduced speed range.
Lastly, a design with high \emph{MUF} is desirable, where \emph{MUF} is defined as the ratio of average torque to PM volume.
A primary screening based on \emph{THDV} of Pareto-optimal solutions shows that the solutions lying in the bottom region of the Pareto-front must be avoided as they have more than 30\% \emph{THDV}, as shown in Figure~\ref{fig:Pareto-THDV}.
Since the remaining Pareto-optimal solutions have similar \emph{THDV} (10-14\%), it is easier to select solutions based on the remaining performance measures. 
Consequently, three preferred solutions, 1, 2, and 3, highlighted in Figure~\ref{fig:Pareto-THDV}, are selected after further evaluation.
The basis of the selection of the solutions is as follows.

\begin{itemize}
    \item Solution 1: maximum average torque
    \item Solution 2: maximum \emph{MUF}
    \item Solution 3: minimum pulsation and \emph{F-BEMF}
\end{itemize}

\subsubsection{Trade-Off Calculation Using Objective Functions}

A trade-off analysis of the Pareto-optimal front is an effective method to select preferred solutions without domain expertise.
In this article, for a particular solution ($\mathbf{x}^{(i)}$), trade-off is calculated using \eqref{eq:trade-off} on the neighborhood of points, represented by $B(\mathbf{x}^{(i)})$.
The term $\sum^{M}_{m=1}\{1|c_m > d_m\}$ calculates the number of $m$'s (out of $M$) for which the condition $c_m > d_m$ is valid.
In this article, the complete Pareto-optimal set obtained from NSGA-II-WR and NSGA-II-WR-SA, as explained previously, defines the neighborhood of points, $B(\mathbf{x}^{(i)})$.
For trade-off calculation, only two objective functions defined in~\eqref{eqn:electric-problem} are used and solutions with high trade-off values are desired.
\begin{equation}\label{eq:trade-off}
\begin{split}
    \texttt{Avg.Loss}(\mathbf{x}^{(i)},\mathbf{x}^{(j)}) &= \frac{\sum^{M}_{m=1}\max\left( 0,f_m(\mathbf{x}^{(j)}) - f_m(\mathbf{x}^{(i)})\right)}{\sum^{M}_{m=1}\{1|f_m(\mathbf{x}^{(j)}) > f_m(\mathbf{x}^{(i)})\}}, \\[2mm]
    \texttt{Avg.Gain}(\mathbf{x}^{(i)},\mathbf{x}^{(j)}) &= \frac{\sum^{M}_{m=1}\max\left( 0,f_m(\mathbf{x}^{(i)}) - f_m(\mathbf{x}^{(j)})\right)}{\sum^{M}_{m=1}\{1|f_m(\mathbf{x}^{(i)}) > f_m(\mathbf{x}^{(j)})\}}, \\[2mm]
    \texttt{Trade-off}(\mathbf{x}^{(i)}) &= \max^{|B(\mathbf{x}^{(i)})|}_{j=1} \frac{\texttt{Avg.Loss}(\mathbf{x}^{(i)},\mathbf{x}^{(j)})}{\texttt{Avg.Gain}(\mathbf{x}^{(i)},\mathbf{x}^{(j)})}.
\end{split}
\end{equation}
After performing the trade-off calculation, three solutions with the highest trade-offs, Solution 3, 4, and 5, are identified from the combined Pareto-optimal set, as shown in Figure~\ref{fig:Pareto-Trade-off}.
The trade-off values of the selected solutions are given below.
Interestingly, Solution 3 is picked again with the highest trade-off value.
Additionally, since Solutions 4 and 5 offer a relatively smaller trade-off value compared to Solution 3, they are not considered in the rest of the discussion.

\begin{itemize}
    \item Solution 3: highest trade-off value (114.99)
    \item Solution 4: $2^{nd}$ highest trade-off value (50.79)
    \item Solution 5: $3^{rd}$ highest trade-off value (35.07)
\end{itemize}

\subsubsection{Performance comparison of selected solutions}

\begin{figure}
\centering
\subfigure[\label{fig:MCDM-PCP}]{
\includegraphics[width=0.48\textwidth]{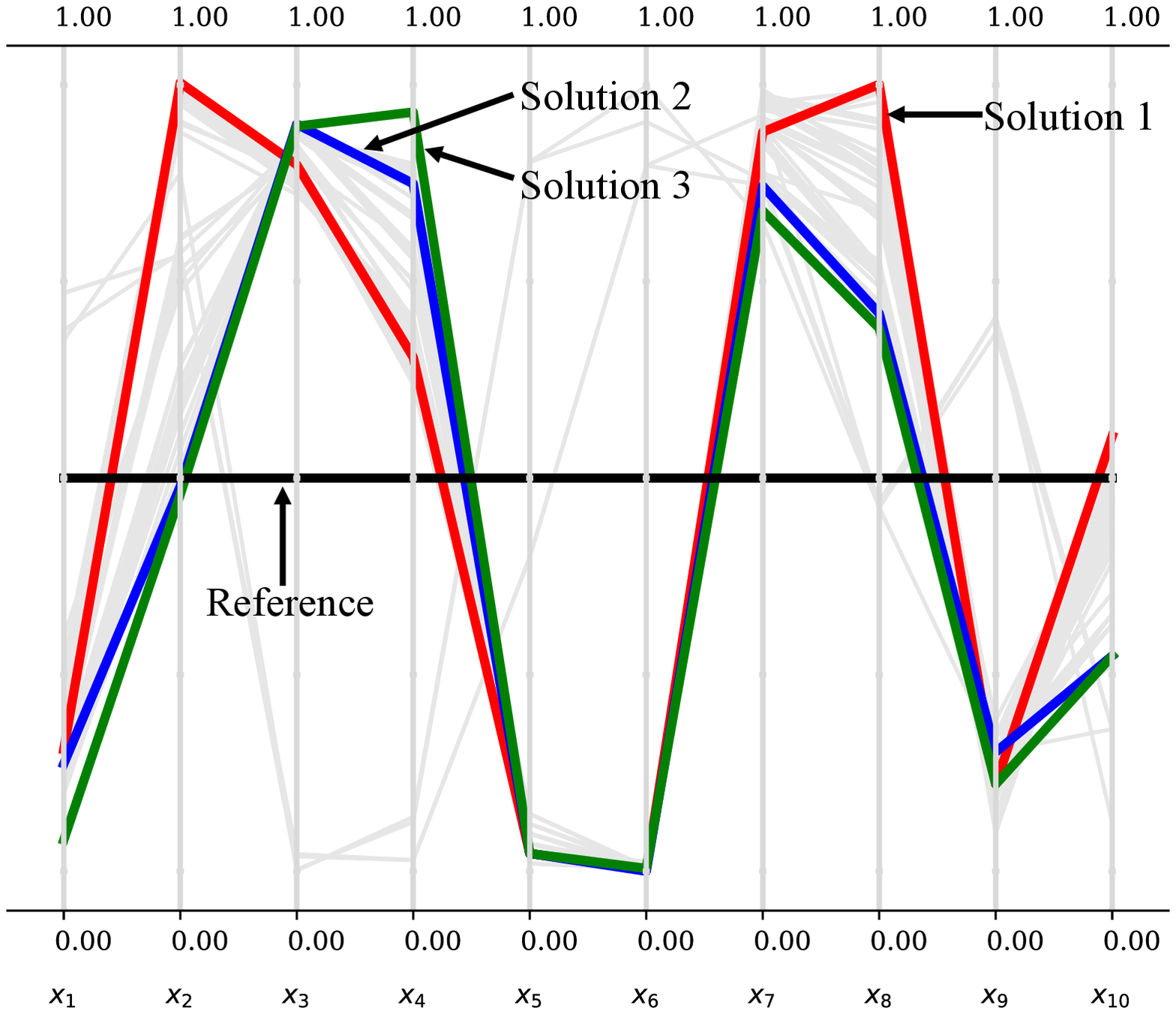}}
\hfill
\subfigure[\label{fig:Torq-speed}]{
\includegraphics[width=0.48\textwidth]{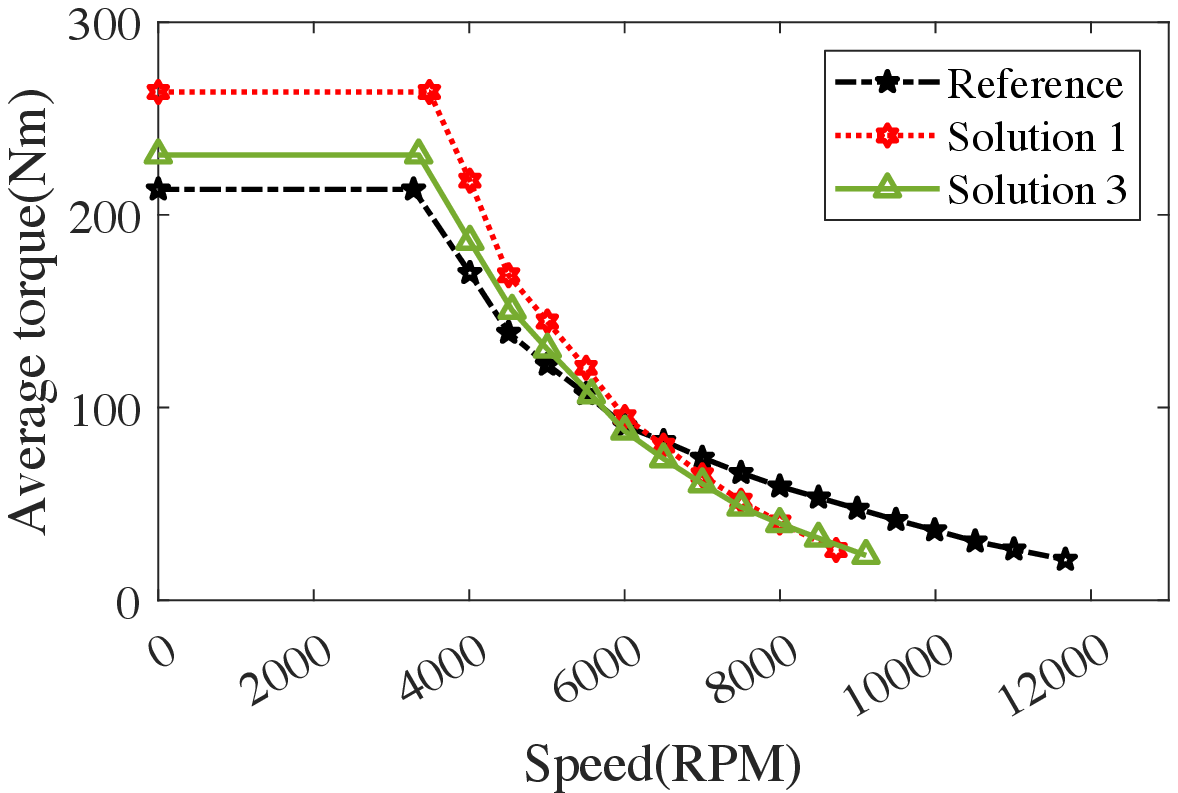}}
\caption{Selected Pareto-optimal solutions are highlighted in normalized design space in \subref{fig:MCDM-PCP}. Torque-speed curve of reference design and solutions 1 and 3 is shown in \subref{fig:Torq-speed}.}
\label{fig:Comparison-MCDM}
\end{figure}

Performance details of the selected solutions and the reference design are given in Table~\ref{tbl:MCDM-details}.
Further insights into the performance of these solutions can be gained by analyzing the design space, as shown in Figure~\ref{fig:MCDM-PCP}.
Some essential observations highlighting the trade-off among selected solutions are as follows.

\begin{table}
\tbl{Performance comparison of five preferred solutions found using domain specific a posteriori MCDM method and trade-off analysis. Preferred values are highlighted in bold for the three solutions.}
{\begin{tabular}{ccccccccccc} \toprule
\multirow{2}{*}{\textbf{Solution}} && \textbf{Avg torque} && \textbf{Pulsations} && \textbf{THDV} && \textbf{MUF} && \textbf{F-BEMF} \\
&& (Nm) && (Nm) && (\%) && (Nm/mm$^3$) && (V) \\
\midrule
1 && $\mathbf{263.0374}$ && 47.4060 && 14.1263 && 0.0290 && 248.2401\\
2 && 235.5986 && 14.2488 && $\mathbf{11.2982}$ && $\mathbf{0.0308}$ && 236.7291\\
3 && 231.3853 && $\mathbf{9.9186}$ && 11.5016 && 0.0304 && $\mathbf{234.4451}$\\
\hline
\multirow{2}{*}{Reference} && \multirow{2}{*}{214.7760} && \multirow{2}{*}{36.1846} && \multirow{2}{*}{14.4093} && \multirow{2}{*}{0.0330} && \multirow{2}{*}{209.2622} \\ \\
\bottomrule
\end{tabular}}
\label{tbl:MCDM-details}
\vspace{-2mm}
\end{table}

\begin{figure}
\centering
\subfigure[Solution 1.\label{fig:Solution1-B}]{
\includegraphics[width=0.4\textwidth, trim= 100 100 40 115, clip]{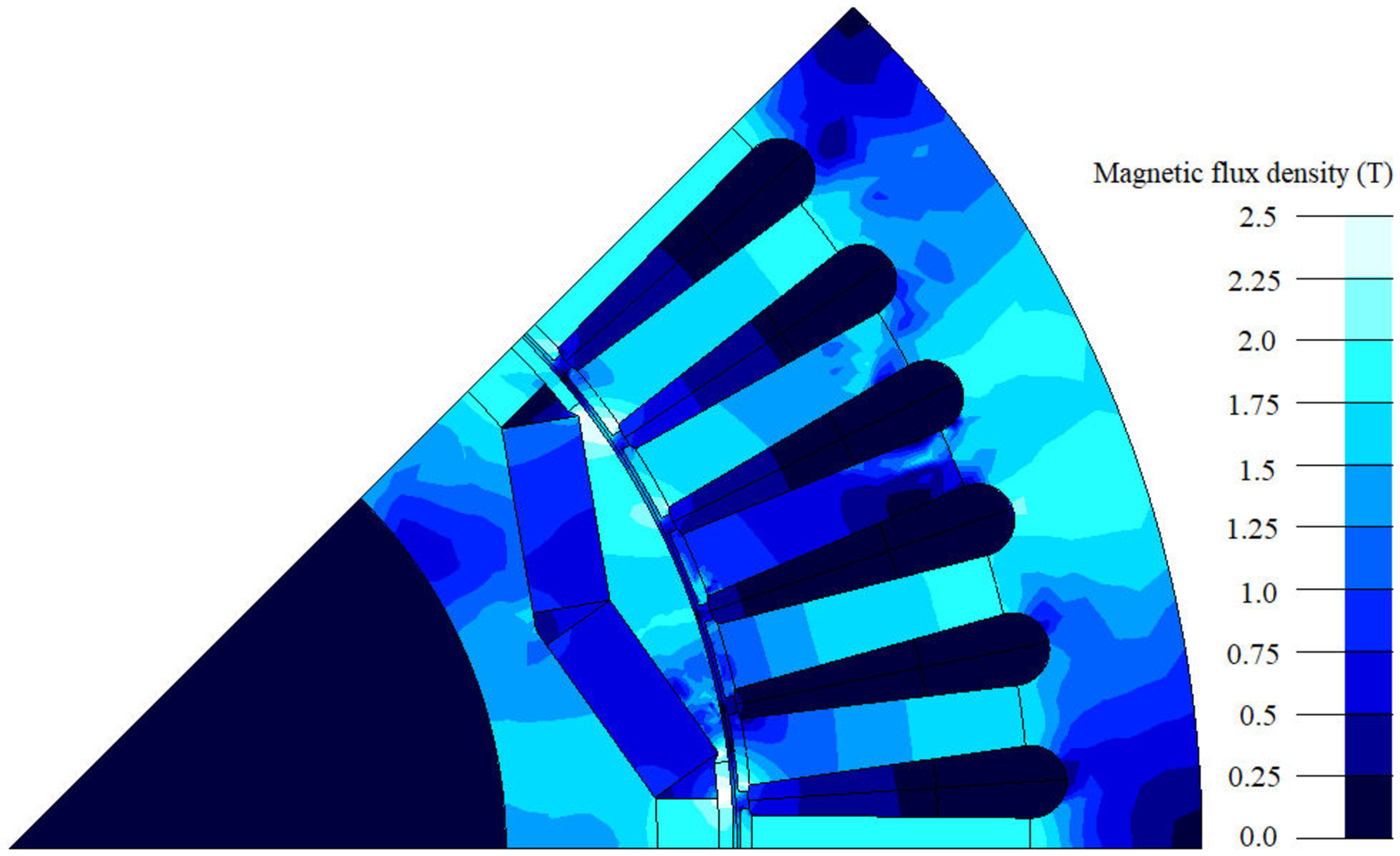}}
\hspace{2mm}
\subfigure[Solution 2.\label{fig:Solution2-B}]{
\includegraphics[width=0.4\textwidth, trim= 100 100 40 115, clip]{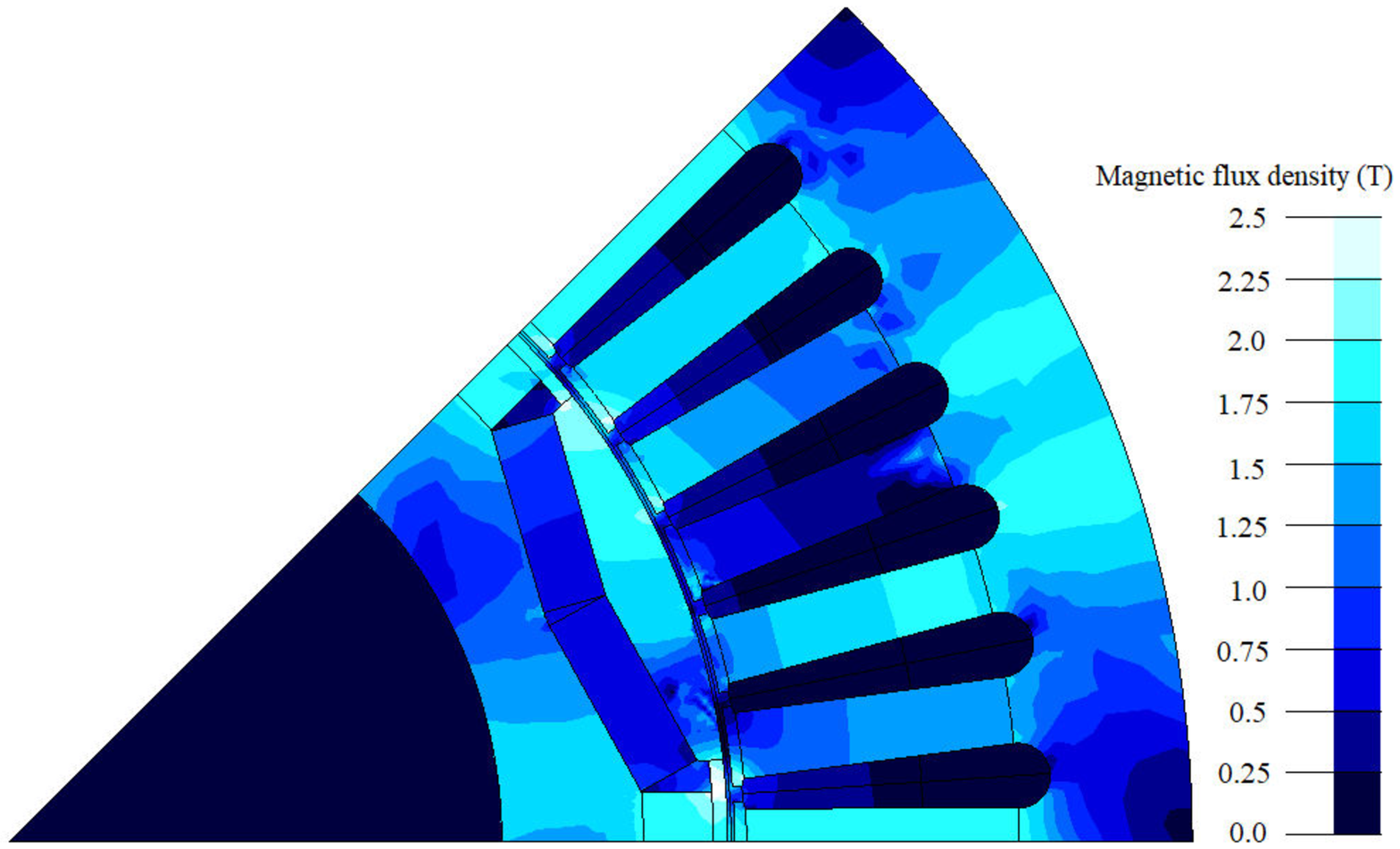}}
\subfigure[Solution 3.\label{fig:Solution3-B}]{
\includegraphics[width=0.4\textwidth, trim= 100 100 40 115, clip]{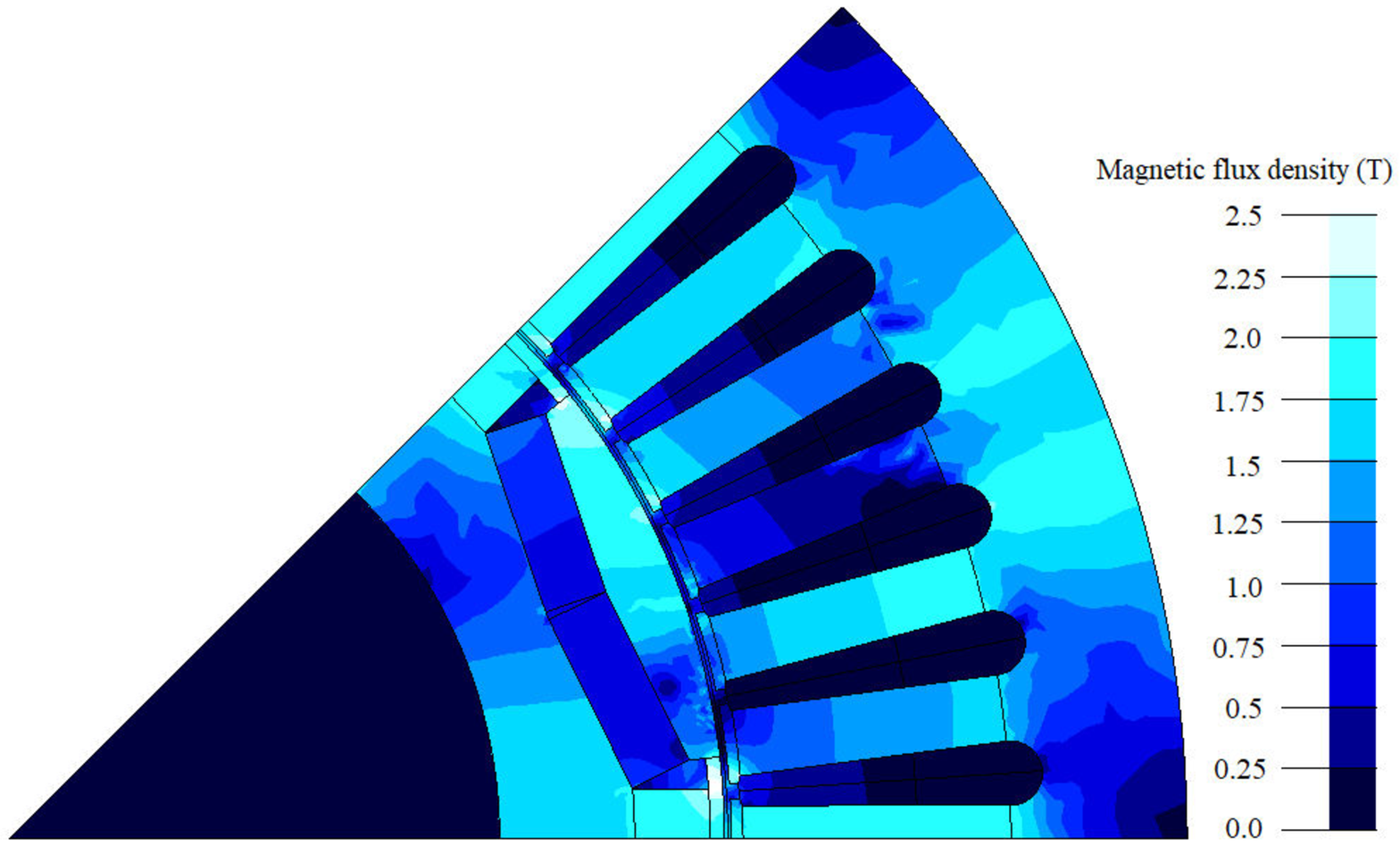}}
\hspace{2mm}
\subfigure[Reference design.\label{fig:Reference-B}]{
\includegraphics[width=0.4\textwidth, trim= 100 100 40 115, clip]{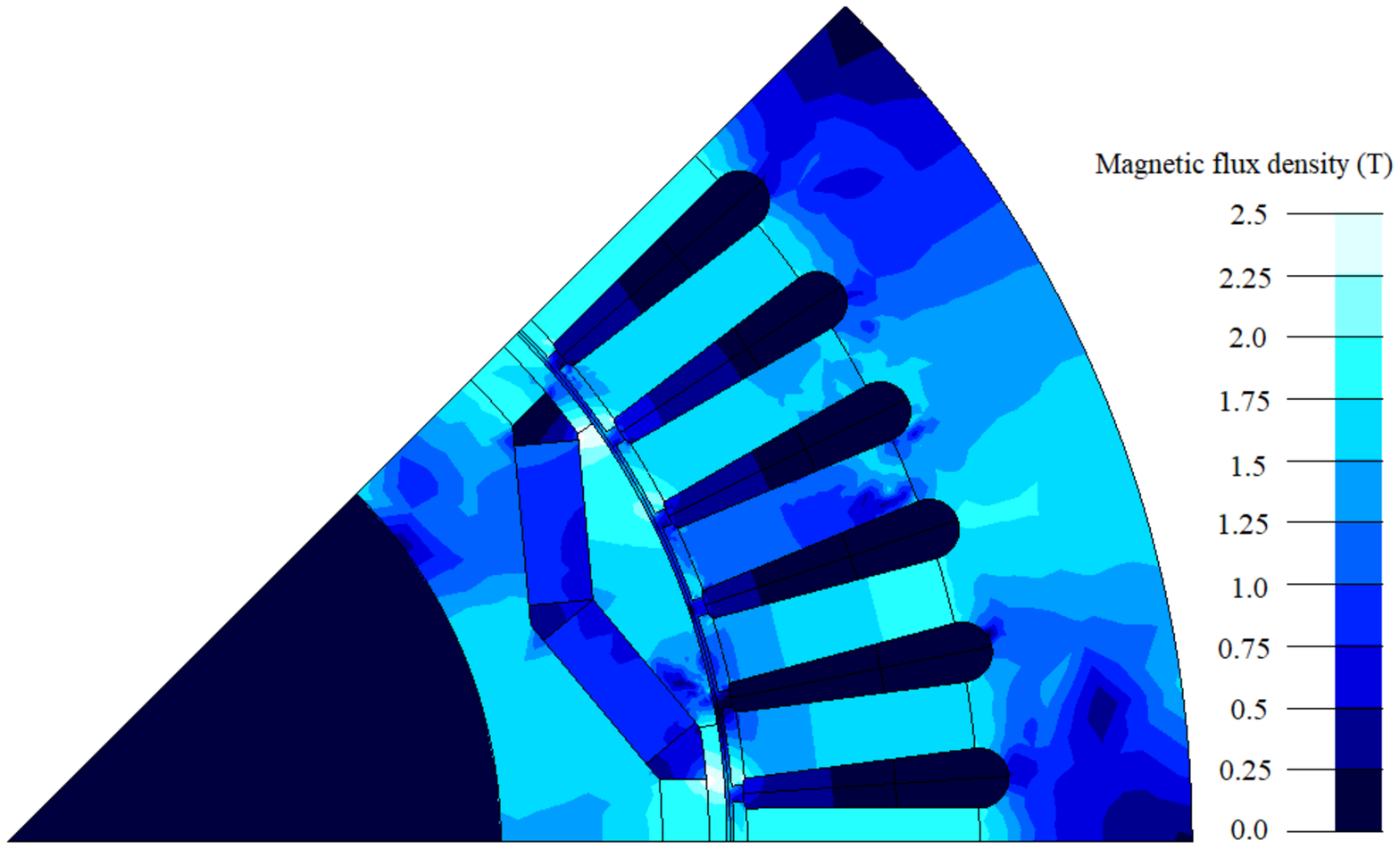}}
\caption{Magnetic flux density plots of Solutions 1, 2, 3 and reference design at rated operation.} 
\label{fig:B-solutions}
\end{figure}

\begin{itemize}
    \item Although Solution 1 provides maximum average torque; it also has the maximum amplitude of pulsations and \emph{F-BEMF}. Both these characteristics can be explained by larger magnet thickness ($x_2$), slot height ($x_7$), slot width ($x_8$), and slot opening height and width ($x_9$ and $x_{10}$).
    \item Solutions 2 and 3 perform quite similarly in all aspects, with slight variations observed in average torque and torque pulsations.
    This variation is caused by the different angles between magnets ($x_4$) observed for the two solutions.
    \item All selected solutions have a larger \emph{F-BEMF} value compared to the reference design. In other words, they have a lower speed range.
    The relation between \emph{F-BEMF} and the maximum achievable speed is illustrated in Figure~\ref{fig:Torq-speed}.  With further increase in speeds, one would observe that torque produced by Solution 1 drops to zero more quickly compared to Solution 3. Since Solutions 2 and 3 have similar \emph{F-BEMF}, their torque/speed profiles are also expected to be similar.
    \item A comparison of the magnetic flux density plots of Solutions 1, 2, and 3 at corresponding rated operating conditions reveals that Solution 1 suffers from higher saturation in stator teeth, back iron, and rotor steel close to magnet edges (see~Figure~\ref{fig:B-solutions}).
\end{itemize}

Based on this in-depth discussion, one should select Solution 1 for an application with a high average torque requirement. 
If the focus is more on a smooth operation with a high-speed range, Solution 2 or 3 should be chosen.
It is also worth mentioning that while the trade-off analysis can pick Solution 3, it does not pick Solution 2 with the highest \emph{MUF}, but can be selected by utilizing domain expertise.
Ultimately, selecting a single solution out of a Pareto-optimal set is a difficult task that can often be alleviated using the machine designer's experience.

\section{Conclusion}

This article has investigated a bi-objective electric machine design optimization problem with geometric constraints.
While the geometric constraints are evaluated using analytical expressions, the objective functions require costly finite element analysis, leading to a mixed computational expensive optimization problem.
The proposed method has utilized a repair operator to handle inexpensive constraints and surrogate models to predict expensive objectives. Both concepts have been integrated into the well-known evolutionary multi-objective optimization (EMO) algorithm NSGA-II.
Results have indicated an improved quality of the Pareto-optimal solution set using the repair operator and thus the effectiveness of ensuring feasibility during optimization.
Moreover, the surrogate incorporation has been analyzed in-depth and shown to be critical for improving efficiency.
First, surrogate related parameters have been investigated by performing a sequential parametric study, examining the number of infill solutions in each generation ($N$), the number of generations for exploiting the surrogate model ($k$), and the number of the initial design of experiments ($N^{\texttt{DOE}}$).
The parametric study has provided a suitable configuration for solving this electric machine design problem.
Results have validated the superiority of incorporating surrogates by improving the algorithm's convergence with a small number of expensive evaluations.

The ultimate goal of an optimization process is to reach an optimal solution that can be implemented successfully. Unlike many other applied multi-objective optimization studies, this article has presented a domain-specific a posteriori MCDM approach focusing on machine cost, noise, vibration, and harshness (NVH), and speed range of the electric machine to choose a single preferred solution. The presented a-posteriori selection approach has identified the trade-off offered by different Pareto-optimal solutions and facilitated selecting a handful of optimized electric machine designs.
Additionally, trade-off calculations based on objective functions have been performed to select preferred solutions, thereby helping the user identify a few critical designs from a large search space.

Future research shall be conducted on an approach to perform parameter tuning automatically.
Moreover, this article has explored the effectiveness of surrogates by analyzing the accuracy of objective value predictions.
It has been observed that nonlinear objective functions may delay the algorithm's convergence. In electric machines, the nonlinear behavior of magnetic material is dependent on the operating
point.
A space-filling sampling technique, which systematically explores the entire objective space, could be an effective way to account for this non-linearity before optimization. However, in real-world optimization problems, such as electric machine design, prior information about objective space is usually not available.
This can be the subject of future research in the electric machine community.
Integrating a repair operator and surrogate models into an EMO algorithm has shown promising results for optimizing electric machine designs.
The proposed concepts' generalizability shall be further explored by incorporating them into other EMO methods. Moreover, more application problems having a computationally mixed expensive nature need to be investigated. Nevertheless, this study has clearly shown the advantage of using a flexible EMO algorithm with efficient handling of constraints and the use of surrogates for expensive evaluation procedures to discover a diverse set of high-performing designs. The study has also revealed a set of key design principles common to multiple high-performing designs for enhancing knowledge about the problem and demonstrated the use of MCDM approaches to choose one or a few preferred solutions for implementation. The complete optimization-cum-decision-making on a complex electric motor design problem demonstrated in this study should pave the way for applying similar procedures in other engineering design optimization tasks.
\section{Data Availability Statement}

The data that support the findings of this study are available from the corresponding author upon reasonable request.
\section{Disclosure Statement}

No potential competing interest was reported by the authors.

\bibliographystyle{tfcad}
\bibliography{references}

\end{document}


\articletype{SUPPLEMENTARY DOCUMENT}

\title{Optimized Electric Machine Design Solutions with Efficient Handling of Constraints and Surrogate Assistance}

\author{
\name{Bhuvan Khoshoo\textsuperscript{a}\thanks{CONTACT Bhuvan Khoshoo. Email: khoshoob@msu.edu}, Julian Blank\textsuperscript{b}, Thang Q. Pham\textsuperscript{a}, Kalyanmoy Deb\textsuperscript{a}, \\and Shanelle N. Foster\textsuperscript{a}}
\affil{\textsuperscript{a}Department of Electrical and Computer Engineering, Michigan State University, East Lansing, USA; \textsuperscript{b}Department of Computer Science and Engineering, Michigan State University, East Lansing, USA}
}

\maketitle

\begin{abstract}
    This supplementary document contains additional details, results, and explanations of ideas presented in the main paper. Constraint formulations and details of repair operator implementation are provided. Additionally, this document contains setup and results of parametric study that explains the effect of surrogate models on the convergence of optimization algorithm.
\end{abstract}

\section{Formulation of Geometric Constraints}

Optimization of electrical machines typically starts with the selection of a machine template.
Based on machine performance requirements, objective functions are defined, followed by the selection of the design variables, the variable ranges, and constraints.
Average torque, torque pulsations, losses, and efficiency are some of the most common objective functions used in optimizing machine design.
Designers usually select variables based on domain knowledge and then proceed to sensitivity analysis to keep only the most significant variables during optimization \citep{2000-Gillon}.
Variable ranges can be selected using a naive approach or based on the machine designer's experience.
Since the search space is generally high dimensional, a proper definition of geometric constraints ensures the reliability of obtained solutions.
Moreover, it is important to define the desired operating point for optimization.

Definition of feasible search space for an optimization problem requires selection of optimization variables, variable ranges, and formulation of constraints.
Some important parameters of the 48-slot/8-pole interior permanent magnet (IPM) machine selected for optimization in this study are given in Table~\ref{tbl:machine-ratings}.
For a fair comparison, following parameters are kept constant as in the original design.

\begin{itemize}
    \item Inner diameter (ID) and outer diameter (OD) of stator and rotor
    \item Air-gap length between stator and rotor
    \item Stack length of the machine
    \item Number of turns per coil
    \item Slot fill factor and maximum current density in stator slot
\end{itemize}

\begin{figure}[hbt]
    \centering
    \includegraphics[width=0.5\textwidth, keepaspectratio]{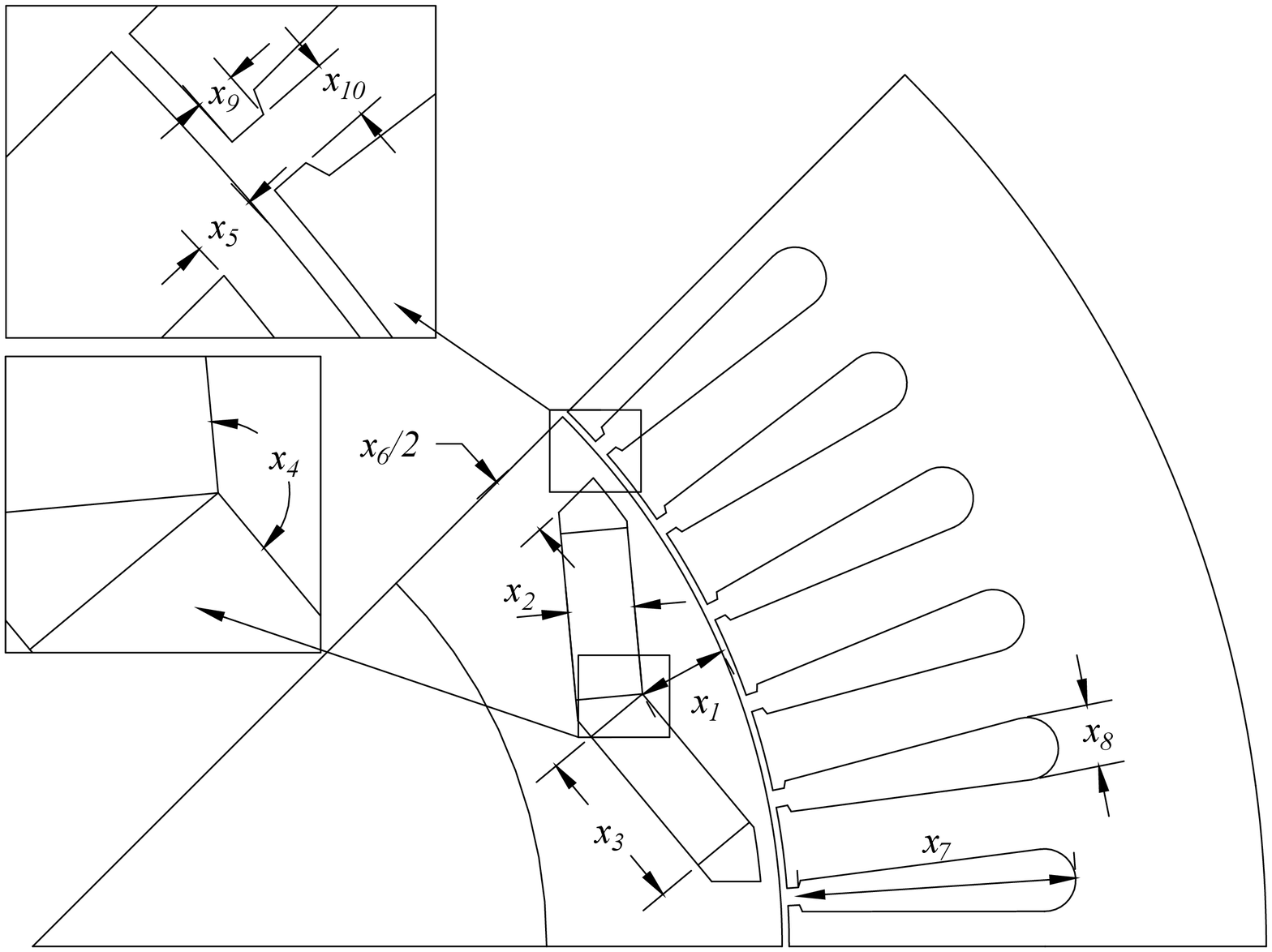}
    \caption{Geometric variables used for optimization.}
    \label{fig:Opt-vars}
\end{figure}

Next, a sensitivity analysis study is performed to identify the ten most significant geometric variables which are also shown in Figure~\ref{fig:Opt-vars}.
Out of these ten variables, six ($x_1$ -- $x_6$) are responsible for magnet shape and placement, while the other four variables ($x_7$ -- $x_{10}$) control the stator slot shape and back iron thickness.
Once the optimization variables are identified, optimization constraints are formulated which ultimately define the feasible search space.
In this work, ten geometric constraints are used which are relatively inexpensive to calculate compared to selected objective functions.
Although geometric constraints are inexpensive, their formulation is not trivial since it involves calculation of other geometric parameters. 
Therefore, in this work, a parametric model is used which ensures that all ten constraints are implemented correctly for each solution.
Exact formulation of these constraints is listed below for reference.
For details of the symbols used, Flux-2D model of the selected IPM machine can be accessed \citep{Altair-fluxmotor}.
Constants used for constraint formulation are given in Table~\ref{tbl:Constants}.
Definition of all other geometric parameters required for evaluation of constraints is given in (\ref{eqn:other-1}) - (\ref{eqn:other-last}).
Ultimately, the ten geometric constraints used in optimization problem are given in (\ref{constraint-g1}) - (\ref{constraint-g10}).
It is worth mentioning that the geometric constraints defined in this article are general and will ensure feasibility of design for any arbitrarily selected variable ranges.

\begin{table}
\tbl{Parameters of IPM machine used for optimization.}
{\begin{tabular}{lcclc}
\toprule
\textbf{Parameters} & \textbf{Values} && \textbf{Parameters} & \textbf{Values} \\
\midrule
Mechanical power & 69$\,$kW && Turns per coil & 11\\
Rated speed & 3000$\,$rpm && Slot/ pole/ phase & 2\\
Peak current & 177$\,$A && Slot fill factor & 0.46\\
Stator outer diameter & 264$\,$mm && Air-gap & 0.75$\,$mm\\
Rotor outer diameter & 160.4$\,$mm && Stack length & 50.8$\,$mm\\
Magnet type & NdFeB && DC-link voltage & 650$\,$V\\ 
\bottomrule
\end{tabular}}
\label{tbl:machine-ratings}
\end{table}

\begin{table}[h!]
\tbl{Constants used for constraint formulation.}
{\begin{tabular}{lcccccc} 
\toprule
Constant & OS\_SN &	IM\_PN & OS\_ID & AG\_L & IM\_ID & ZIM\_UV\\
Value & 48 & 8 & 161.9 & 0.75 & 110 & 1 \\ 
\midrule
Constant & IM\_V1 & IM\_W2 & OS\_WS1 & OS\_H1 & ZOS\_UV \\
Value & 0 & 0.7 & 3.34 & 0.27 & 1\\
\bottomrule
\end{tabular}}
\label{tbl:Constants}
\vspace{-6mm}
\end{table}

\begin{equation} \label{eqn:other-1}
    IM\_OD = OS\_ID - 2AG\_L  
\end{equation}
\begin{equation}    
    ZIM\_VE = \frac{180*ZIM\_UV}{IM\_PN}
\end{equation}    
\begin{equation}    
    ZIM\_OR = IM\_OD/2
\end{equation}    
\begin{equation}    
    ZOS\_Y3 = OS\_WS1/2 
\end{equation}   
\begin{equation}    
    ZOS\_VE = \frac{180*ZOS\_UV}{OS\_SN}
\end{equation}    
\begin{equation}    
    ZIM\_X2 = ZIM\_OR - x_1
\end{equation}    
\begin{equation}    
    ZOS\_V1A = x_7 - \frac{x_8}{2} - x_9 - OS\_H1
\end{equation}
\begin{equation}
    ZOS\_V1B = \sqrt{(ZOS\_V1A)^2 - \left( \left(\frac{x_8}{2}\right)^2 - (ZOS\_Y3)^2 \right)}
\end{equation}
\begin{equation}
    ZIM\_X1 = ZIM\_X2 - \frac{x_2}{sin\left((x_4/2)*(\pi/180)\right)}
\end{equation}
\begin{equation}
    ZIM\_Y1 = 0
\end{equation}
\begin{equation}
    ZIM\_X8 = \frac{(ZIM\_OR - x_5)^2 - (x_3 + IM\_W2)^2 + (ZIM\_X2)^2}{2*ZIM\_X2}
\end{equation}
\begin{equation}
    ZIM\_Y8 = \sqrt{(ZIM\_OR - x_5)^2 - (ZIM\_X8)^2}
\end{equation}
\begin{equation}
    ZIM\_X4 = ZIM\_X2 + (x_3*cos\left((x_4/2)*(\pi/180)\right))
\end{equation}
\begin{equation}
    ZIM\_Y4 = x_3*sin\left((x_4/2)*(\pi/180)\right)
\end{equation}
\begin{equation}
    ZIM\_X5 = ZIM\_X4 - (x_2*sin\left((x_4/2)*(\pi/180)\right))
\end{equation}
\begin{equation}
    ZIM\_Y5 = ZIM\_Y4 + (x_2*cos\left((x_4/2)*(\pi/180)\right))
\end{equation}
\begin{multline}
    ZIM\_K1 = ZIM\_Y5 + \frac{x_6}{2*cos\left((ZIM\_VE)*(\pi/180)\right)}\\
    - ZIM\_X5*tan\left(\left(\frac{x_4}{2}+IM\_V1\right)*(\pi/180)\right)
\end{multline}
\begin{equation}
    ZIM\_K2 = tan((ZIM\_VE)*(\pi/180)) - tan\left(\left(\frac{x_4}{2}+IM\_V1\right)*(\pi/180)\right)
\end{equation}
\begin{equation}
    ZIM\_X7  = \frac{ZIM\_K1}{ZIM\_K2}
\end{equation}
\begin{multline}
        ZIM\_Y7 = (ZIM\_X7*tan((ZIM\_VE)*(\pi/180)))\\ 
        - \frac{x_6}{2*cos\left((ZIM\_VE)*(\pi/180)\right)}
\end{multline}
\begin{equation}\label{eqn:other-last}
    ZOS\_V1 = \left(2*atan2\left(ZOS\_V1A-ZOS\_V1B, \frac{x_8}{2}+ZOS\_Y3 \right)\right)*(180/\pi)
\end{equation}

\begin{equation} \label{constraint-g1}
    g_1 = -\left(\sqrt{(ZIM\_X1)^2 + (ZIM\_Y1)^2} - \left(\frac{IM\_ID}{2}+2\right)\right)
\end{equation}
\begin{equation} \label{constraint-g2}
    g_2 = \sqrt{(ZIM\_X4)^2 + (ZIM\_Y4)^2} - \left(\frac{IM\_OD}{2}-1\right)
\end{equation}
\begin{equation} \label{constraint-g3}
    g_3 = -(ZIM\_X7 - ZIM\_X5)
\end{equation}
\begin{equation} \label{constraint-g4}
    g_4 = -(ZIM\_Y7 - ZIM\_Y5)
\end{equation}
\begin{equation} \label{constraint-g5}
    g_5 = \sqrt{(ZIM\_X7)^2 + (ZIM\_Y7)^2} - \left(\frac{IM\_OD}{2} - x_5 - 1\right)
\end{equation}
\begin{equation} \label{constraint-g6}
    g_6 = ZIM\_X8 - ZIM\_X4 - 3
\end{equation}
\begin{equation} \label{constraint-g7}
    g_7 = ZIM\_Y8 - ZIM\_Y4 - 1
\end{equation}
\begin{equation} \label{constraint-g8}
    g_8 = -(OS\_WS1 - x_{10})
\end{equation}
\begin{equation} \label{constraint-g9}
    g_9 = |ZOS\_V1 - ZOS\_VE| - ZOS\_VE
\end{equation}
\begin{equation} \label{constraint-g10}
    g_{10} = -\left(\frac{ZIM\_X7 - ZIM\_X5}{(x_4/2)*(\pi/180)} - 0.1\right)
\end{equation}

\section{Selection of Operating Point For Optimization}

\begin{figure}
    \centering
    \includegraphics[width=0.7\textwidth, keepaspectratio]{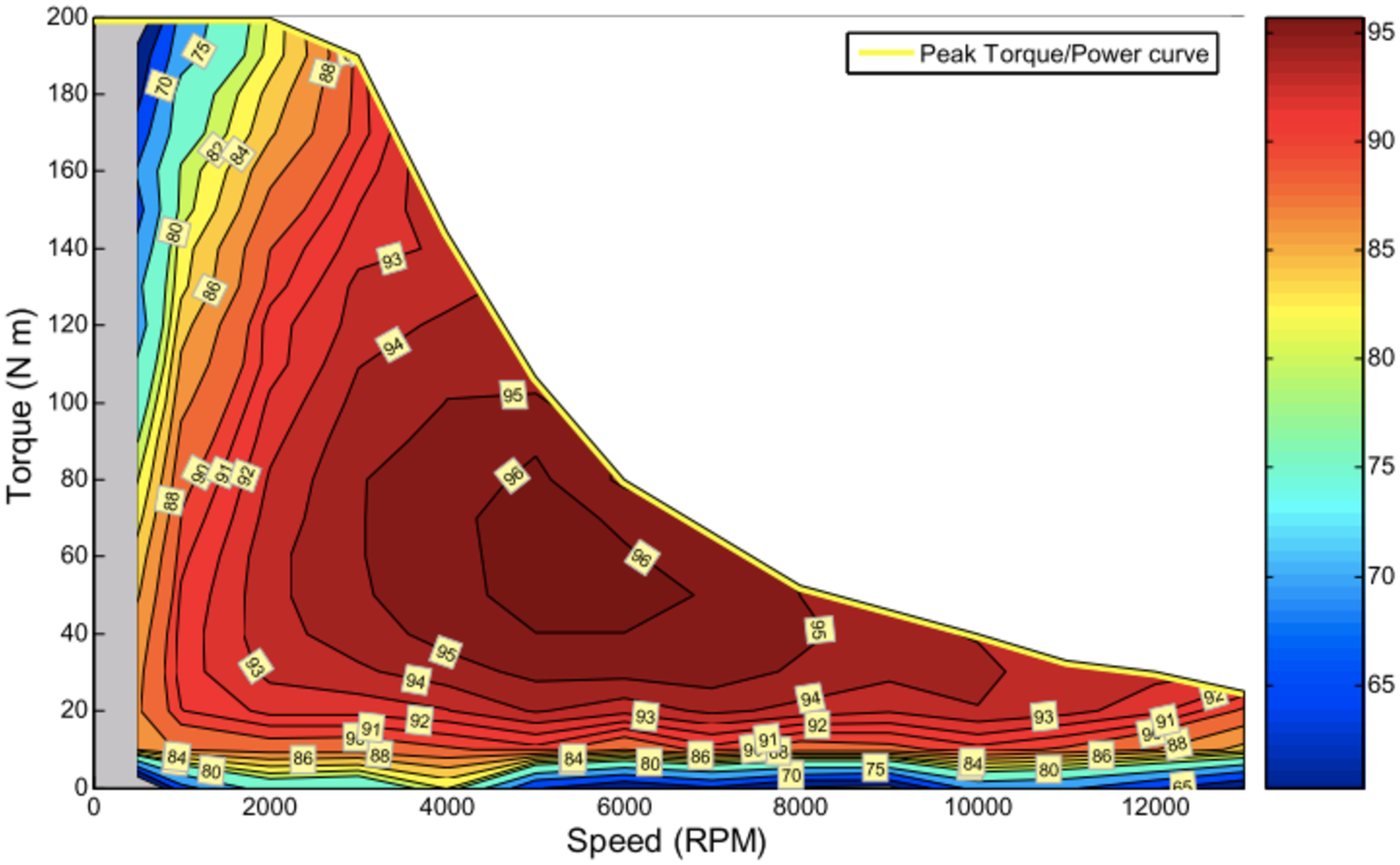}
    \caption{ \label{fig:Prius-eff} 2010 Prius motor efficiency contours for 650 Vdc \citep{ORNL-Prius2010}.}
\end{figure}

The performance of an IPM machine directly depends on the speed and torque requirements.
The efficiency contour map of the 2010 Prius motor with a dc-link voltage of 650$\,$V is shown in Figure~\ref{fig:Prius-eff} \citep{ORNL-Prius2010}. 
The peak torque rating of the machine stays constant till a particular speed, called the base speed of the machine, which is also the rated speed of the machine.
PMSMs are famous for their torque density, and therefore, their operation at rated speed is of high importance.
To increase the efficiency of PMSMs in the speed region up to the base speed, they are operated in a way to minimize the excitation current fed to stator copper windings and, therefore Joule losses, while meeting the torque requirements.
This specific mode of operation is called maximum torque per ampere (MTPA) operation.
In this work, the rotational speed of the rotor and the excitation angle is kept constant to those of reference design at rated MTPA operation.
It should be noted that the excitation current is not kept constant during optimization. The slot fill factor denotes the proportion of slot area filled by copper windings. As the slot fill factor is kept constant in contrast to the slot cross-section, it results in different current ratings for different designs.

\section{Implementation of Repair Operator}

A recommended property of the repair procedure is to find a feasible solution close to the original infeasible solution but at the intersection of the infeasible constraints.
Additionally, the optimization variables must respect the manufacturing accuracy limitations, which limits them to have only two decimal places.
To achieve both these objectives, the repair operator is implemented in following two phases.
It should be noted that in this work, the repair operator is combined with evolutionary multi-objective optimization (EMO) algorithm NSGA-II~\citep{Deb-NSGA2}.
NSGA-II is a modular, parameter-less optimization algorithm well suited for bi-objective optimization problems, including optimization of electric machines. NSGA-II starts with a population of random solutions called the parent population. After evaluating the population members, pair-wise comparisons are made to select non-dominated and less-crowded solutions~\citep{2001-deb-moo} to meet the main goals of multi-objective optimization. The selected population members are then recombined and mutated to create an offspring population of the same size as the parent population. After evaluating the offspring population, it is merged with the parent population to execute a final survival selection to pick the top half of the population. The selected population becomes the parent population of the next generation. This process is continued until a termination criterion is satisfied.
Although this article uses NSGA-II as the base optimization algorithm, other EMO methods can also be tried as long as the constraints used in optimization problem formulation are inexpensive to calculate.

\subsection{Geometric Constraint Repair} 
Given an infeasible solution $\bx$, the goal of this phase is to convert it to a feasible solution $\bx'$.
For this purpose an optimization problem is solved which is defined in \eqref{eqn:repair}.
Both the geometric constraints and box constraints ensure that the resulting solution ($\bx'$) of \eqref{eqn:repair}  is feasible, whereas the objective $\left\Vert \bx - \bx' \right\Vert^2$ favors the feasible solution having the closest Euclidean Distance (in $\ell_2$-norm) to the original solution~$\bx$.

\begin{equation}
\begin{aligned}
\minimize_{\bx'}~  & \; \left\Vert \bx - \bx' \right\Vert^2\\[0.5mm]
\text{subject~to}~ & \quad g_j(\bx') \leq 0 & \forall j\in 1,\ldots,10~~\\
& \quad x_i^{(L)} \leq x'_i \leq x_i^{(U)}, & \forall i\in 1,\ldots,10~\\[1mm]
\text{where}  & \quad \bx' \in \mathbb{R}^N.
\end{aligned}
\label{eqn:repair}
\end{equation}

In this study, the above optimization problem is solved by the gradient-free simplex optimization algorithm proposed by Nelder and Mead~\citep{1965-nelder-mead}. The algorithm requires an initial solution set to $\bx$, thereby searching is focused and quick. The output of the first phase is $\bx'$, which is a feasible real-valued vector.

\subsection{Precision Repair} 
After the first phase of the repair has been completed, every variable $x'_i$ needs to be modified to make it a floating-point number with a precision of two. 
This can be achieved by rounding each variable $x'_i$ ($1\leq i\leq N$) to its floor or ceil value to the desired precision. 
Because such rounding will result in $2^N$ different possibilities, an efficient rounding scheme is proposed.   
 
The method starts by creating a random permutation $P$ of the first $N$ consecutive natural numbers ($|P| = N$). The permutation $P$ provides the order of the variables to be rounded sequentially. 
After that, a local search method inspired by Hooke-Jeeves pattern moves~\citep{hooke-jeeves} is implemented to find a feasible solution having two-decimal precision in all variables. 
For the current variable under consideration, two changes (its floor and ceil) are evaluated for the entire solution's feasibility. 
By keeping the rounding that caused feasibility, the subsequent variable in $P$ is checked. 
A new random permutation is then tried if the process does not find a feasible solution after all variables are explored. 
If a solution $\bx$ cannot be repaired in a maximum of $\rho$ ($=100$ used here) attempts, it is discarded.
This is an efficient way of finding a feasible solution out of all $2^N$ neighbouring possibilities.


\begin{figure}[hbt]
    \centering
    \includegraphics[width=0.5\textwidth, keepaspectratio, trim=0.3 10.3cm 19cm 0,clip]{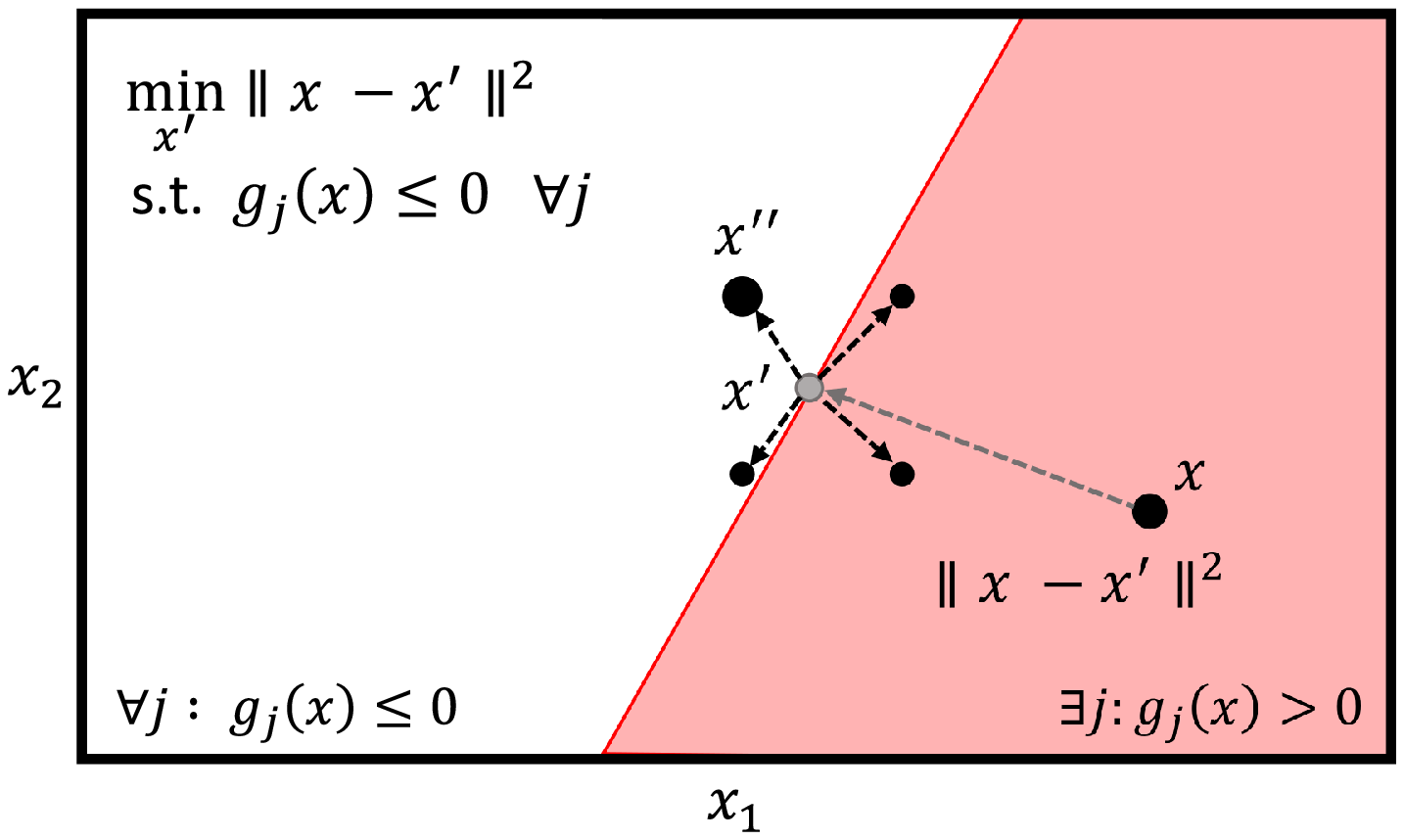}
    \caption{Illustration of the repair operator in a 2-D  search space.}
    \label{fig:repair-example}
\end{figure}

An illustration of the repair operator and its two phases is shown in Figure~\ref{fig:repair-example}. The two-dimensional search space is split up into two regions, the feasible one where for $\forall j \in J: g_j(\bx) \leq 0$ and infeasible one where  $\exists j \in J: g_j(\bx) > 0$.
A solution $\bx$ which is located in the infeasible area, the repair operator attempts to make it feasible. 
The first phase executes the embedded optimization method obtaining $\bx'$, which is feasible and minimizes the Euclidean distance to $\bx$.
In the second phase, $\bx'$ is modified to be a float of precision two by rounding. 
In this example, in a two-dimensional search space ($N=2$), the rounding can result in $2^N = 2^2 = 4$ different solutions where 50\% turn out to be still feasible. 
Nevertheless, as soon as the first attempt of randomly ordered rounding is successful, the repair is completed, and $\bx''$ is returned. 
In all possible cases, the above two phases can produce feasible solutions with two-decimal places of rounding given that the constraints are inexpensive to evaluate.

\section{Analysis of Constraints}

To analyze the impact of geometric constraints formulated for the optimization problem, an initial study is conducted where 100 solutions are sampled randomly using Latin Hypercube sampling. To ensure statistical significance, the experiment is repeated 100 times. Results show that $69.7\%$ of these randomly sampled solutions are infeasible, or in other words, only $30.3\%$ are feasible. A solution is considered infeasible if at least one constraint is violated. Thus, an analysis of each constraint separately shall provide more information on what type of constraints are more difficult to satisfy.

In Table~\ref{tbl:constraints}, the percentage of infeasible solutions is shown for each constraint separately. The percentages reveal that some constraints seem to be more difficult to satisfy than others. The constraints $g_8$ and $g_9$, which control the slot shape, and $g_1$, which controls the magnet placement close to the shaft, have not been violated in any of the $10,000$ solutions. In contrast, some other constraints such as $g_2$ and $g_7$, related to magnet placement close to rotor OD, are responsible for infeasible solutions 31.43\% and 36.28\% of the time.
It is worth mentioning that these $10,000$ solutions were generated from the variables ranges defined for the optimization problem studied in this article.
The percentage of infeasible solutions is likely to be larger for broader variable ranges.

\begin{table}
\tbl{The constraint violation of each constraint value from $g_1$ to $g_{10}$.}
{\begin{tabular}{lcccccccccc} 
\toprule
Value & $g_1$ &	$g_2$ & $g_3$ & $g_4$ & $g_5$ & $g_6$ &	$g_7$ & $g_8$ & $g_9$ & $g_{10}$\\ 
\midrule
Infeasible & 0.0\% & 31.43\% & 19.16\% & 19.16\% & 21.94\% & 19.59\% & 36.28\% & 0.0\% & 0.0\% & 20.25\% \\
Rank & 7 &	2 & 6 & 6 & 3 & 5 &	1 & 7 & 7 & 4\\ 
\bottomrule
\end{tabular}}
\label{tbl:constraints}
\end{table}

\section{Impact of Repair Operator}

\begin{figure}
\centering
\subfigure[Run with median HV (1500 evals).\label{fig:Median-pareto}]{
\includegraphics[width=0.4\textwidth, trim=0 0 20 10, clip]{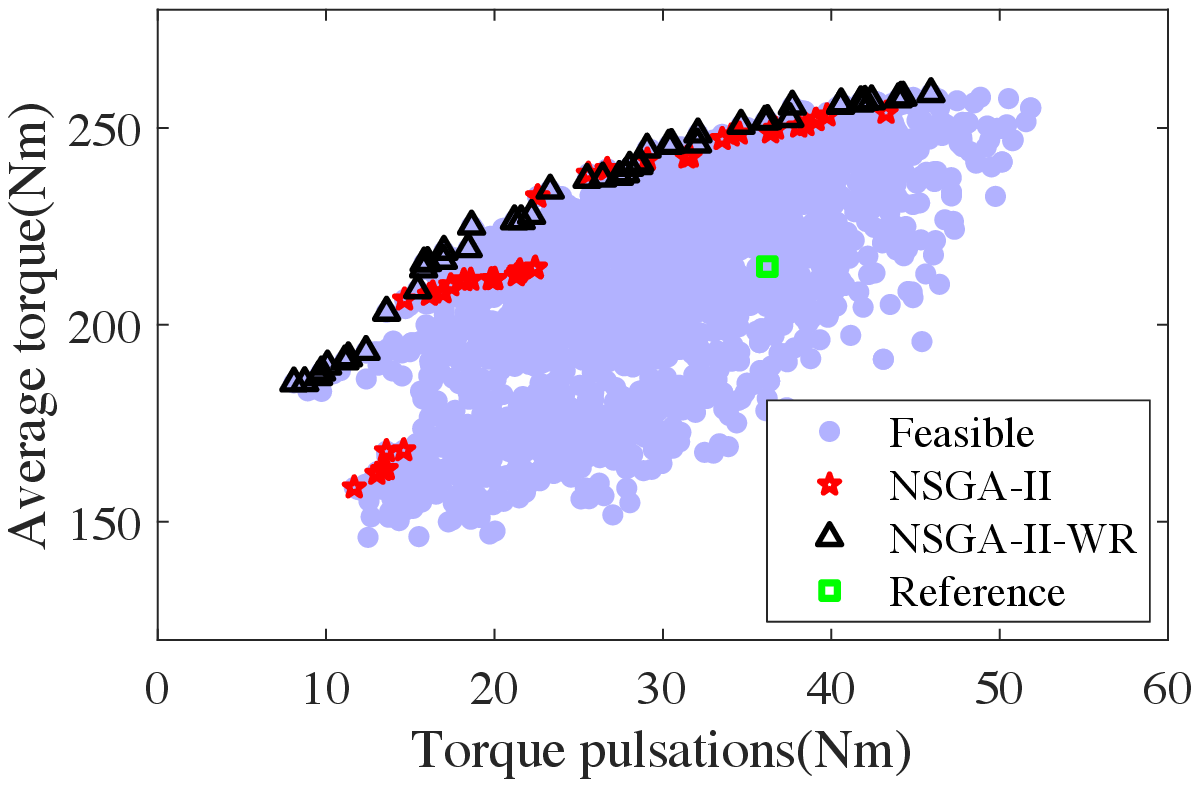}}
\subfigure[Run with the best HV (1500 evals).\label{fig:Best-pareto}]{
\includegraphics[width=0.4\textwidth, trim=0 0 20 10, clip]{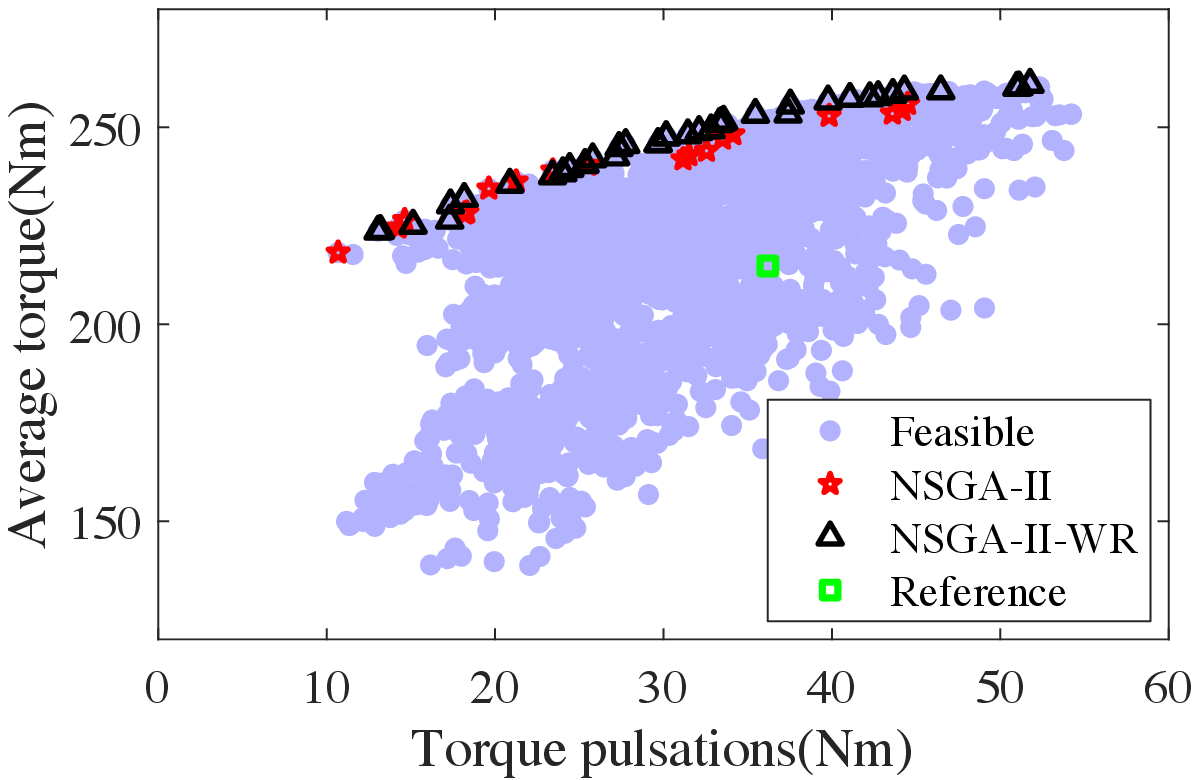}}
\caption{Pareto-optimal fronts of runs with median HV in \subref{fig:Median-pareto} and the best HV in \subref{fig:Best-pareto}. All feasible points obtained in the respective runs are shown in blue color.}
\label{fig:Best-and-Median}
\end{figure}

Electric machine design optimization is a computationally expensive problem and users might choose to solve the problem only once if the resources are limited.
Since evolutionary multi-objective algorithms are usually stochastic in nature, it is important to measure the impact of the repair operator on an individual run.
For this purpose, two runs out of the five completed with each method, NSGA-II and NSGA-II-WR, are selected: (1) the run with the median HV, and (2) the run with the best HV, and the corresponding Pareto-optimal fronts are shown in Figure~\ref{fig:Best-and-Median}.
One can observe from runs with median HV that in the absence of the repair operator, the algorithm (NSGA-II) converges to sub-optimal regions with discontinuities, as shown in Figure~\ref{fig:Median-pareto}.
The pareto-optimal front for the run with the best HV obtained with NSGA-II is also discontinuous in contrast to the one obtained by NSGA-II-WR (see~Figure~\ref{fig:Best-pareto}).
Ultimately, the results indicate that the proposed customization is well-suited for optimizing the design of an IPM machine and improves the quality of obtained Pareto-optimal solutions.

\section{Parameter Study for Surrogate-Assisted Optimization}

Although surrogate-assisted optimization is known to find the Pareto-optimal front quicker than other optimization methods, it is also known to be sensitive to (model and optimization-related) hyperparameters.
In this study, the following three hyperparameters are varied to analyze the performance of the proposed optimization method with surrogates.

\begin{itemize}
    \item $N$: Number of \glspl{ese} in each iteration
    \item $k$: Number of surrogate optimization generations for exploitation
    \item $N^{\texttt{DOE}}$: Number of initial design of experiments
\end{itemize}

To carefully analyze the impact of the parameters, three optimization setups (A, B, and C) are defined in a sequential manner.
Each setup consists of four parameter configurations with variations applied to one hyperparameter at a time.
Each configuration is repeated five times with 200 solution evaluations in each run, making the total number of evaluations equal to 1,000.
To make fair comparisons, the same set of $N^{\texttt{DOE}}$ is used for all configurations defined in Setup A and B, while the same seed is used to generate the initial population for all configurations defined in Setup C.
The complete experiment for this parametric study is shown in Table~\ref{tbl:Surrogate-setup}.
Clearly, Setups A, B, and C quantify the impact of $N$, $k$, and $N^{\texttt{DOE}}$, respectively.
To compare the performance of surrogates for different parameter configurations, three criteria are selected: (1) the number of non-dominated solutions ($N_{nds}$), (2) HV, and (3) the rate of change of HV with evaluations (\emph{RHVE}).
This study aims to find the best hyperparameter setting out of all configurations defined in Table~\ref{tbl:Surrogate-setup} in a sequential manner.
Therefore, results of Setup A are used to define Setup B and combined results of Setups A and B are used to define Setup C.
The calculation of the three criteria is explained below.

\begin{itemize}
    \item $N_{nds}$ –- All five runs of each configuration are combined to obtain one Pareto-optimal front, yielding the number of non-dominated solutions. 
    \item HV –- Best and the worst objective function values are found from the Pareto-optimal sets of the configurations being analyzed and objective functions values are normalized to calculate corresponding HV.
    \item \emph{RHVE} –- Five runs of each configuration provide five arrays of \emph{RHVE}. Then median of these five arrays is used to obtain the final \emph{RHVE} for corresponding configuration.
\end{itemize}

\begin{table}
\tbl{Complete setup for analyzing impact of hyperparameters on performance of surrogate-assisted optimization. For all configurations, number of functional evaluations is limited to 200 in a single run. Each configuration is repeated 5 times.}
{\begin{tabular}{cccccccccccccc} \toprule
\multirow{3}{*}{Configuration} & \multirow{3}{*}{Runs} && \multicolumn{3}{c}{\textbf{Setup A}} && \multicolumn{3}{c}{\textbf{Setup B}} && \multicolumn{3}{c}{\textbf{Setup C}} \\
&&& \multirow{2}{*}{$N$} & \multirow{2}{*}{$k$} & \multirow{2}{*}{$N^{\texttt{DOE}}$} && 
\multirow{2}{*}{$N$} & \multirow{2}{*}{$k$} & \multirow{2}{*}{$N^{\texttt{DOE}}$} && 
\multirow{2}{*}{$N$} & \multirow{2}{*}{$k$} & \multirow{2}{*}{$N^{\texttt{DOE}}$} \\ \\
\midrule
1 & 5 && 5 & 25 & 100 && 10 & 10 & 100 && 10 & 35 & 60 \\
2 & 5 && 10 & 25 & 100 && 10 & 20 & 100 && 10 & 35 & 80 \\
3 & 5 && 20 & 25 & 100 && 10 & 25 & 100 && 10 & 35 & 100 \\
4 & 5 && 25 & 25 & 100 && 10 & 35 & 100 && 10 & 35 & 120 \\
\bottomrule
\end{tabular}}
\label{tbl:Surrogate-setup}
\end{table}

\begin{table}
\tbl{Results for Setups A, B, and C defined for analyzing impact of hyperparameters on performance of surrogates. HV is calculated after normalization of objective functions.}
{\begin{tabular}{cccccccccc} \toprule
\multirow{2}{*}{Configuration} && \multicolumn{2}{c}{\textbf{Setup A}} && \multicolumn{2}{c}{\textbf{Setup B}} && \multicolumn{2}{c}{\textbf{Setup C}} \\
&& $N_{nds}$ & HV && $N_{nds}$ & HV && $N_{nds}$ & HV \\
\midrule
1 && 42 & 0.8051 && 32 & 0.8062 && 47 & $\mathbf{0.8650}$ \\
2 && 40 & $\mathbf{0.8304}$ && 33 & 0.7851 && $\mathbf{51}$ & 0.8269 \\
3 && $\mathbf{56}$ & 0.7711 && 40 & 0.8304 && 43 & 0.8211 \\
4 && 30 & 0.7475 && $\mathbf{43}$ & $\mathbf{0.8397}$ && 38 & 0.7983 \\
\bottomrule
\end{tabular}}
\label{tbl:results-setup}
\vspace{-4mm}
\end{table}

\begin{figure}[t]
\centering
\subfigure[\label{fig:NDF-setup-A}]{
\includegraphics[width=0.4\textwidth, trim=0 0 20 10, clip]{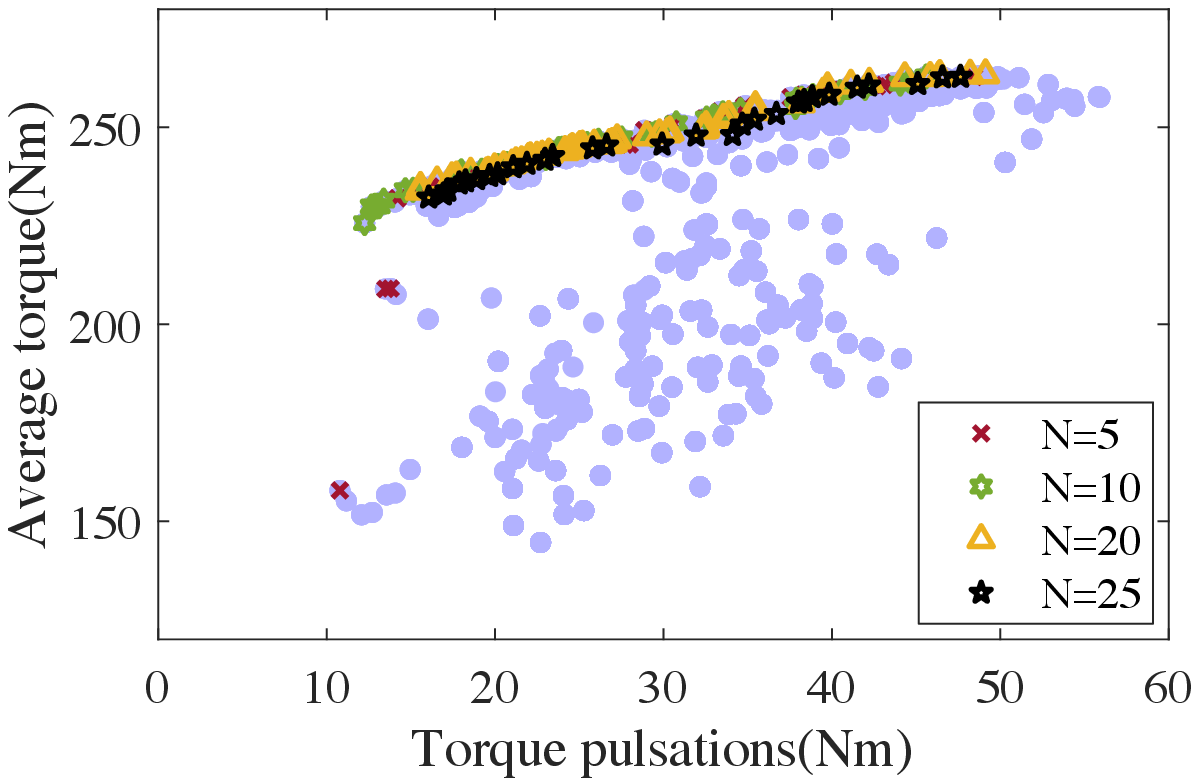}}
\subfigure[\label{fig:HV-setup-A}]{
\includegraphics[width=0.4\textwidth, trim=0 0 20 10, clip]{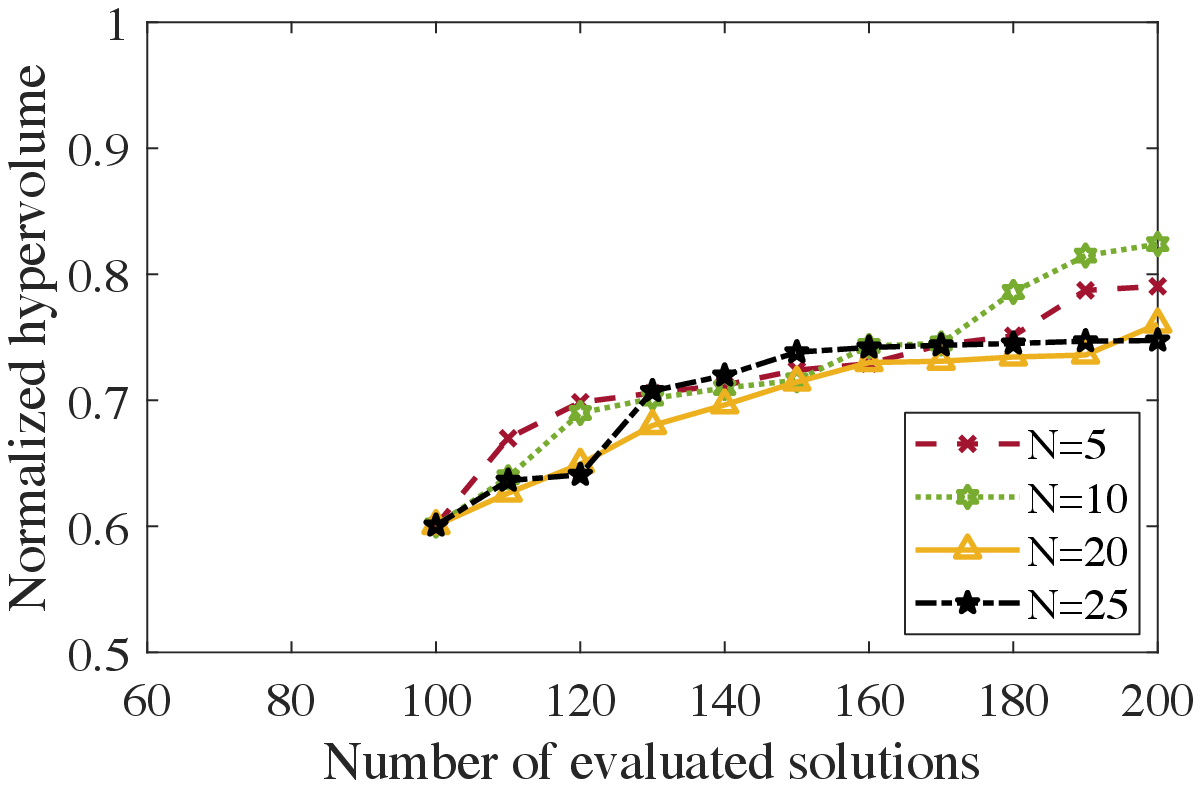}}
\subfigure[\label{fig:NDF-setup-B}]{
\includegraphics[width=0.4\textwidth, trim=0 0 20 10, clip]{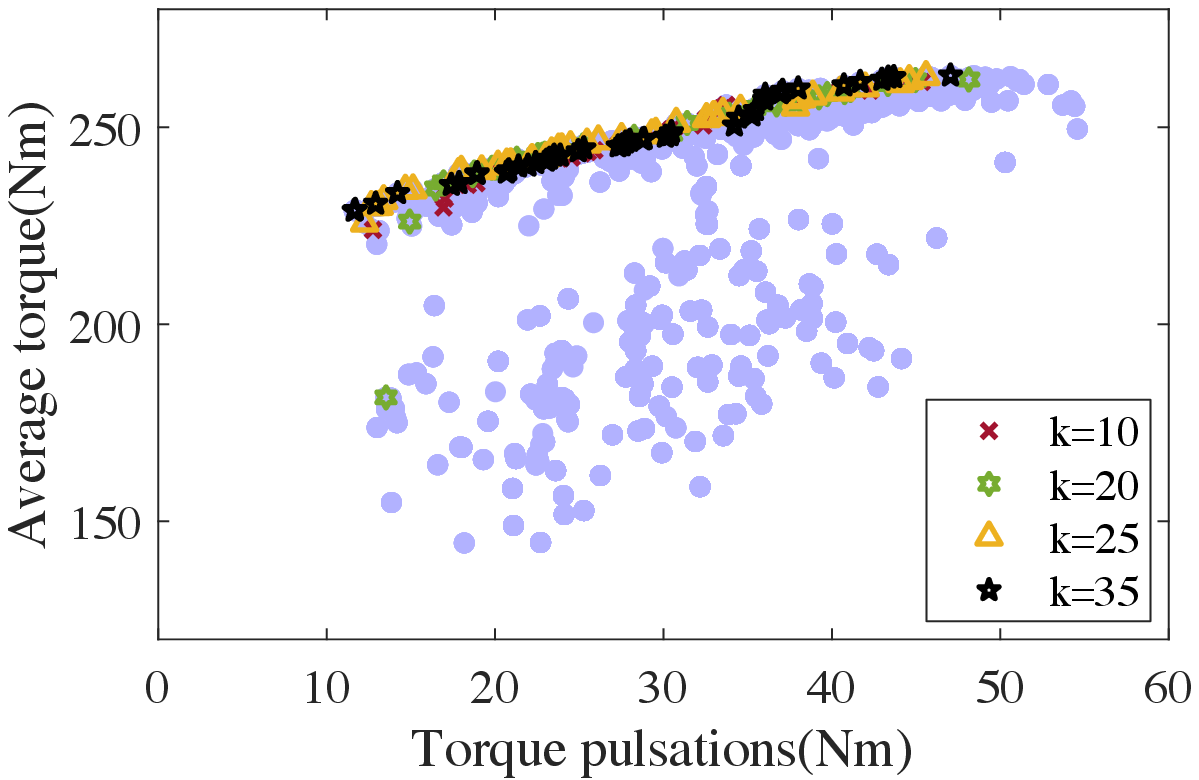}}
\subfigure[\label{fig:HV-setup-B}]{
\includegraphics[width=0.4\textwidth, trim=0 0 20 10, clip]{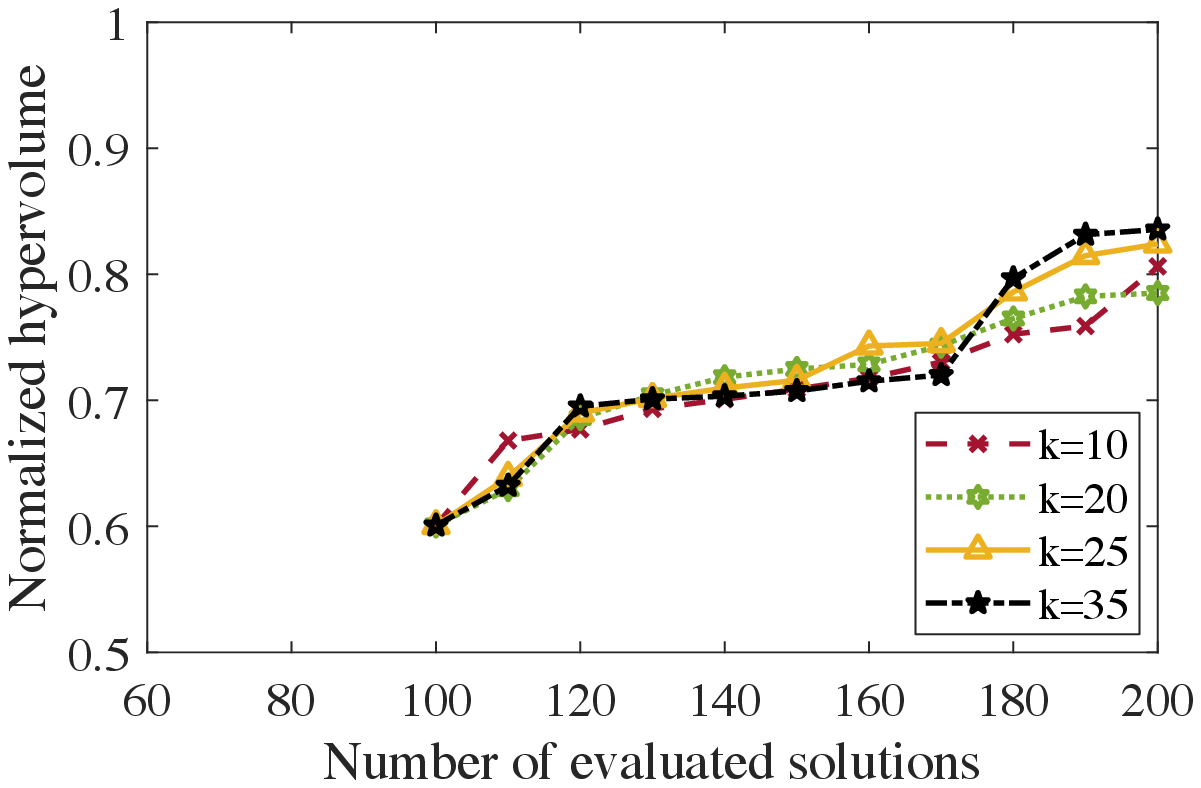}}
\subfigure[\label{fig:NDF-setup-C}]{
\includegraphics[width=0.4\textwidth, trim=0 0 20 10, clip]{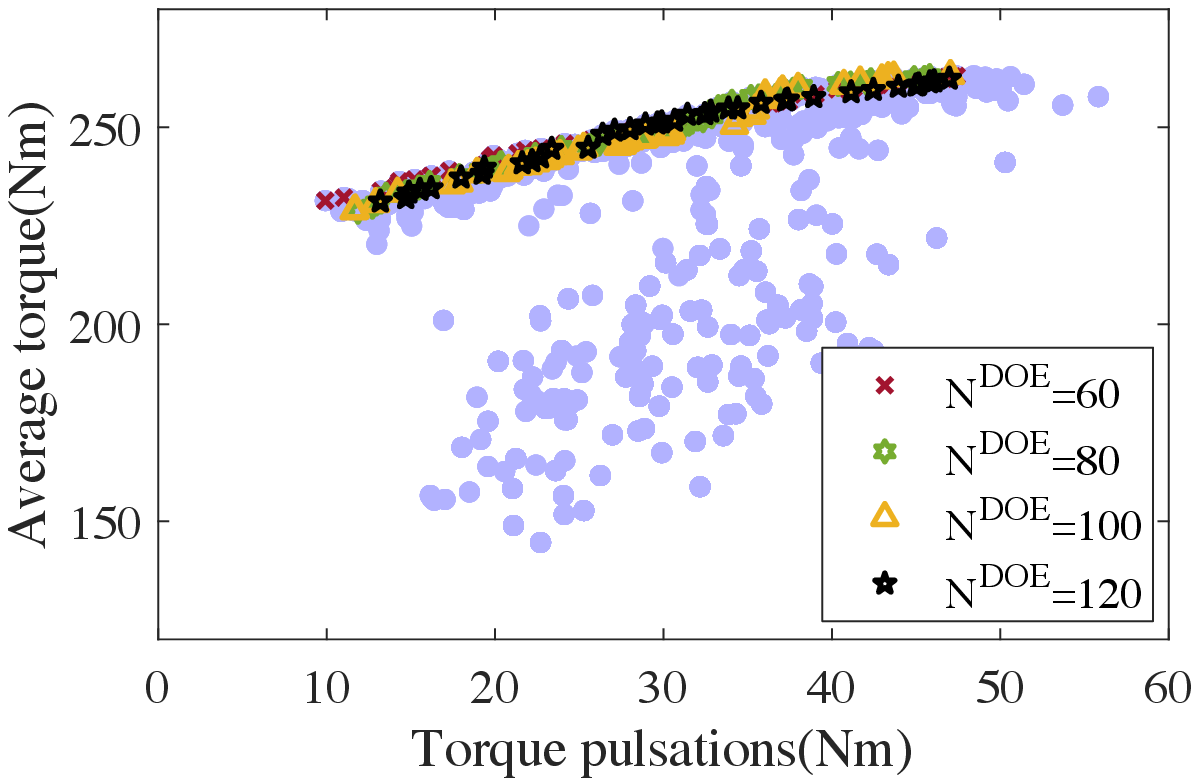}}
\subfigure[\label{fig:HV-setup-C}]{
\includegraphics[width=0.4\textwidth, trim=0 0 20 10, clip]{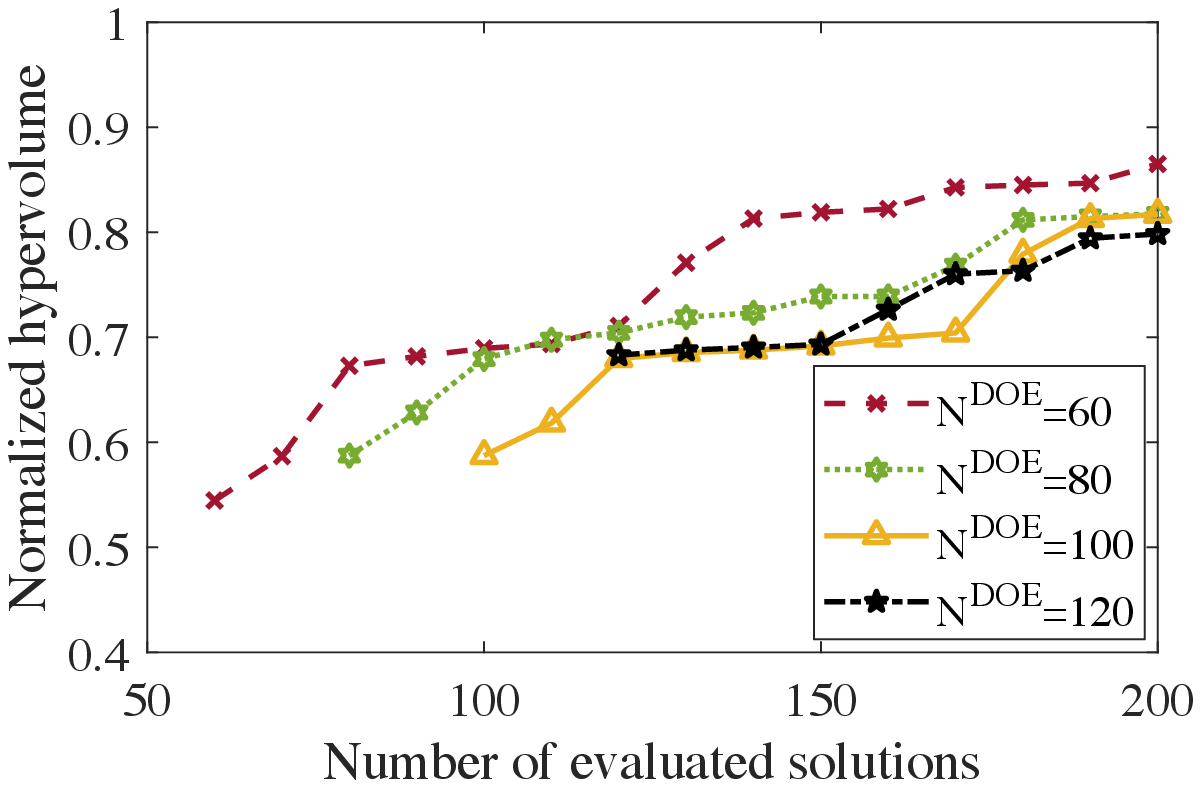}}
\caption{Objective space illustrating Pareto-optimal fronts for configurations of Setups A, B, and C is shown in \subref{fig:NDF-setup-A}, \subref{fig:NDF-setup-B}, and \subref{fig:NDF-setup-C} respectively. \emph{RHVE} for Setups A, B, and C is shown in \subref{fig:HV-setup-A}, \subref{fig:HV-setup-B}, and \subref{fig:HV-setup-C} respectively.} 
\label{fig:results-setupABC}
\end{figure}

Results for all parameter configurations of each setup are shown in Table~\ref{tbl:results-setup} and Figure~\ref{fig:results-setupABC}.
The observations from this study are listed below.

\begin{itemize}
    \item Increasing the number of $N$, initially improves the non-dominated front; however, a large value of $N$ can lead to an over-early convergence with biased search, as can be seen for $N$ = 20 and 25.
    A possible way to overcome this could be to increase the total number of evaluations, which would also result in an unwanted increase in the associated computational cost.
    \item Increasing the value of parameter $k$ leads to a better Pareto-optimal front.
    This makes sense since more generations for exploitation means more number of evaluations before surrogates produce infills with increased accuracy in predictions.
    \item Increasing the value of $N^{\texttt{DOE}}$ results in smaller HV. 
    For a smaller value of $N^{\texttt{DOE}}$, the surrogates have more number of evaluations to improve the model fit and generate better offspring for future generations when the total evaluations are kept constant.
    This means that running the optimization cycle for more evaluations could lead to an improvement in the quality of the Pareto-optimal set for a larger value of $N^{\texttt{DOE}}$.
    However, it will also increase the computational cost of optimization, which is again undesirable. At the same time, one should be careful while reducing the value of $N^{\texttt{DOE}}$ as it could lead to poor model fit and delayed convergence. 
\end{itemize}

\section{Convergence Analysis With and Without Surrogates}

\begin{figure}
\centering
\subfigure[\label{fig:Surr-median}]{
\includegraphics[width=0.4\textwidth, trim=0 0 20 10, clip]{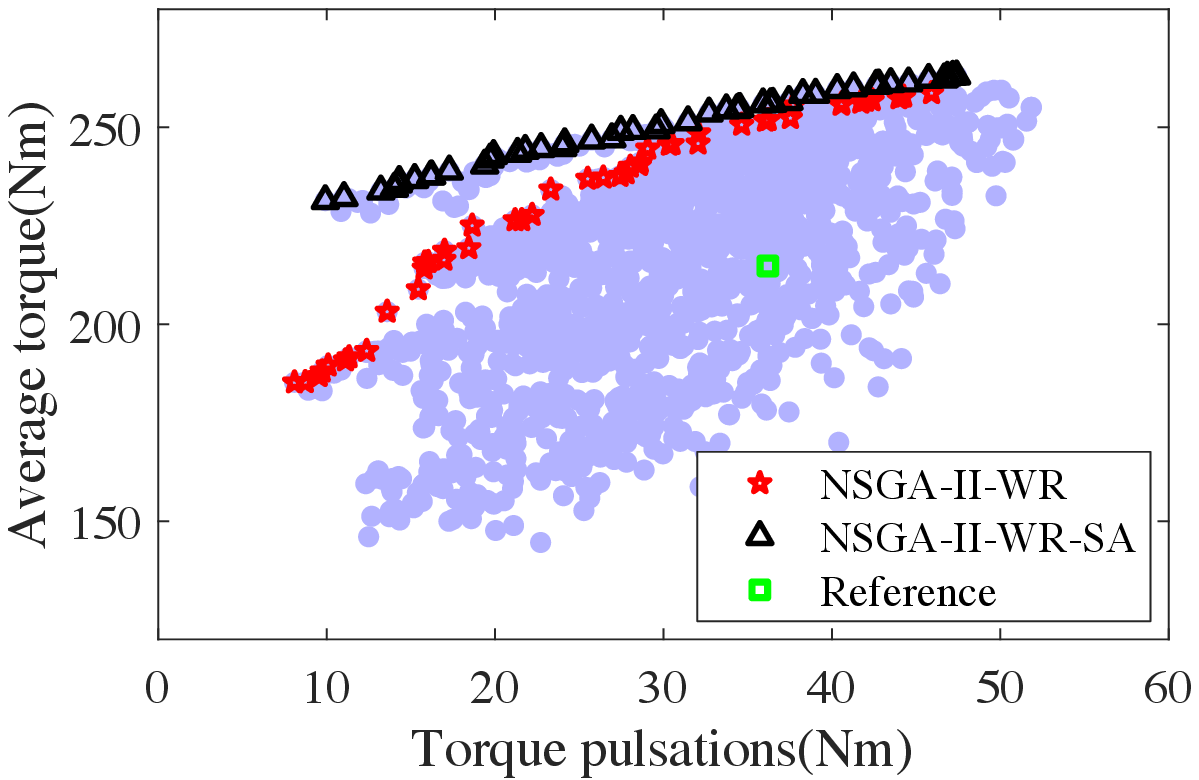}}
\subfigure[\label{fig:Surr-best}]{
\includegraphics[width=0.4\textwidth, trim=0 0 20 10, clip]{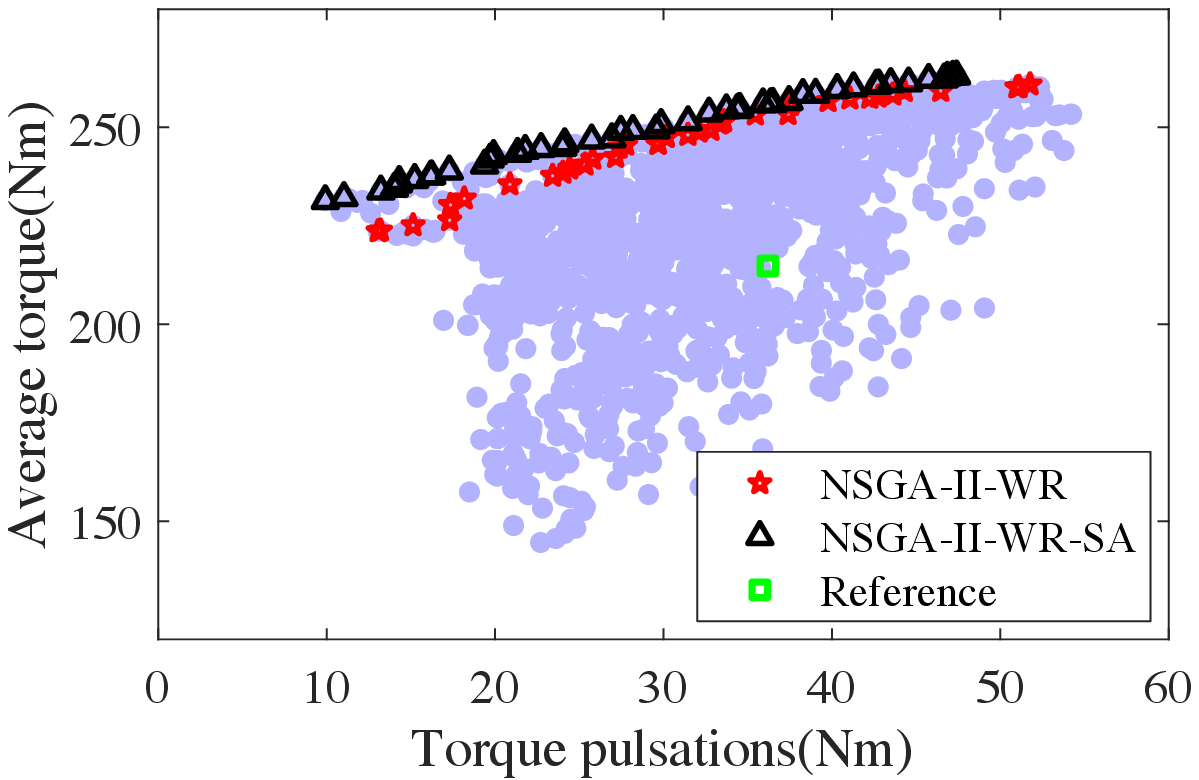}}
\caption{Pareto-optimal fronts of runs with median HV in \subref{fig:Surr-median} and the best HV in \subref{fig:Surr-best}.}
\label{fig:Surr-median-and-best}
\vspace{-2mm}
\end{figure}

It is natural to question the impact of surrogates on an individual optimization run due to the stochastic nature of evolutionary algorithms.
For this purpose, two runs out of the five completed with each method, NSGA-II-WR and NSGA-II-WR-SA, are selected: (1) the run with the median HV, and (2) the run with the best HV, and the corresponding Pareto-optimal fronts are shown in Figure~\ref{fig:Surr-median-and-best}.
Clearly, NSGA-II-WR-SA outperforms NSGA-II-WR in both the compared scenarios.
The difference in the convergence is quite distinct when the runs with median HV are compared as shown in Figure~\ref{fig:Surr-median}.
In fact, it is worth mentioning that NSGA-II-WR-SA was able to reach the same Pareto-optimal front in each of the five runs.
Additionally, a single run of NSGA-II-WR-SA only consisted of 200 expensive evaluations compared to 1500 evaluations in a single run with NSGA-II-WR, which illustrates the impact of surrogates on the algorithm's convergence.

\begin{figure}
\centering
\subfigure[\label{fig:Surrogate-gen}]{
\includegraphics[width=0.4\textwidth]{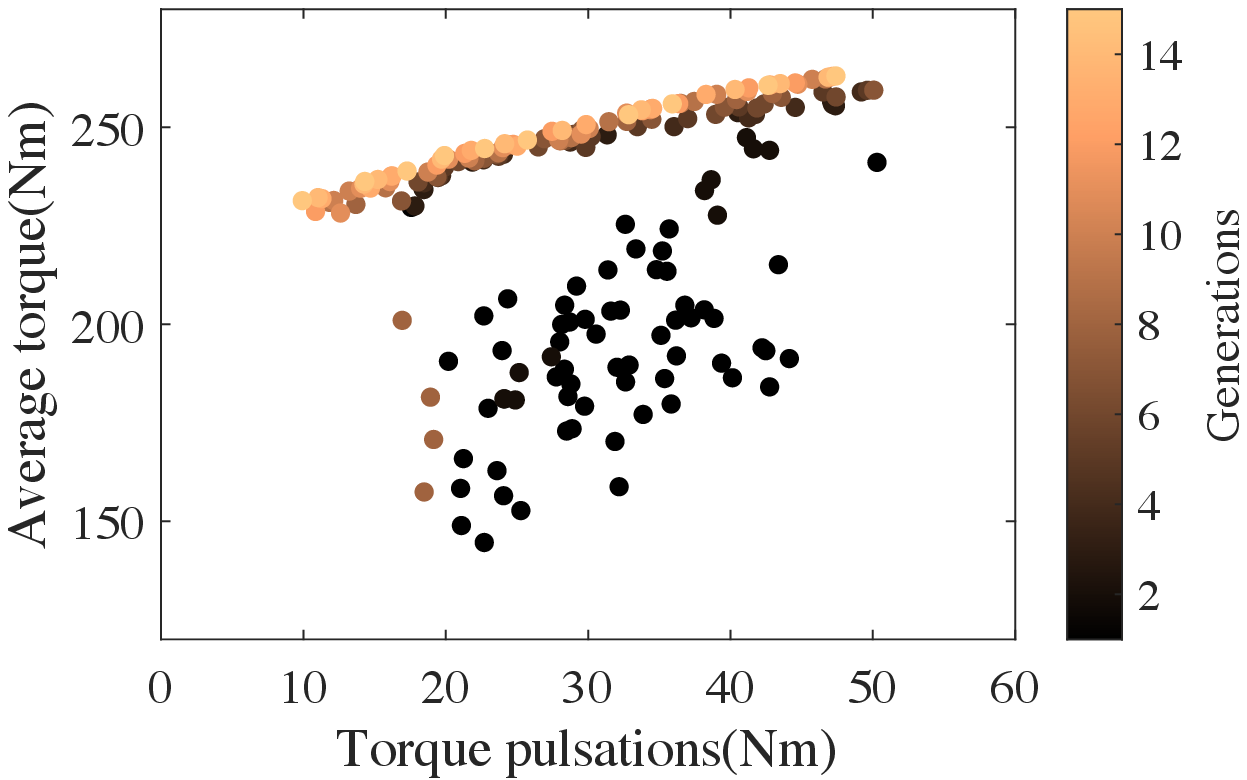}}
\subfigure[\label{fig:Surrogate-MSE}]{
\includegraphics[width=0.4\textwidth]{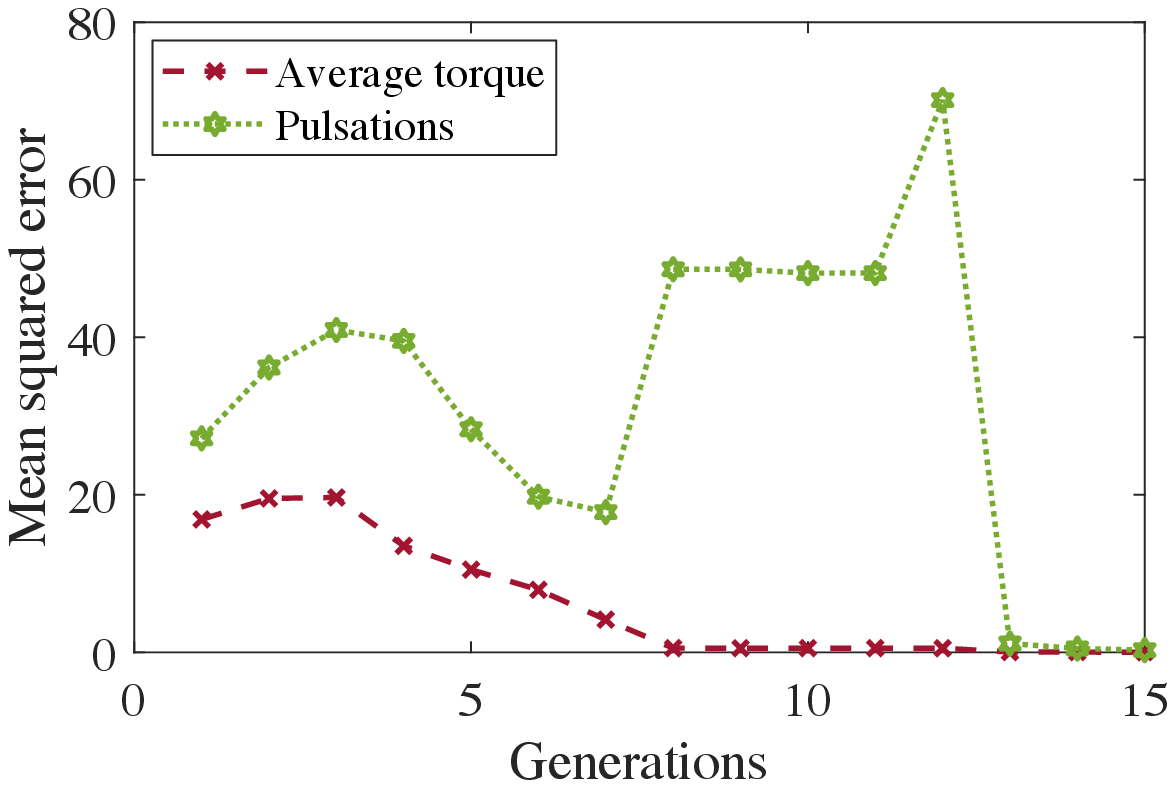}}
\caption{Exploration of objective space and MSE in prediction of objective functions, for each generation using NSGA-II-WR-SA is shown in \subref{fig:Surrogate-gen} and \subref{fig:Surrogate-MSE}, respectively. }
\label{fig:Surrogate-analysis}
\end{figure}

Further insights into convergence can be obtained by analyzing how the surrogates explore the objective space and how accurate the objective values can be predicted.
Figure~\ref{fig:Surrogate-analysis} shows objective space and mean squared error (MSE) of the prediction of objective functions for each optimization cycle of one run (out of five) of NSGA-II-WR-SA.
An important observation is that surrogates can predict average torque with higher accuracy compared to torque pulsations which are greatly influenced by the nonlinearity (magnetic saturation) of electrical steel.
The sudden rise in MSE of pulsations from generation 7 to 8 can be explained by analyzing the corresponding solutions in objective space.
Due to a sudden increase in average torque, more and more solutions are generated with higher slot cross-section and magnet volume, which results in operation in the high saturation region.
Consequently, it takes some time before the surrogates can predict the pulsations accurately.

\section{Conclusions}
In addition to discussions and results of the main paper, this supplementary document has provided further details related to constraint formulation, selection criteria for choosing the operating point for optimization, and the implementation details of the repair operator. Additionally, a sequential parametric study has provided insights into the convergence of algorithm with surrogates. Both documents provide a clear practical surrogate-based optimization methodology for handling applied problems having multiple conflicting objectives in finding a set of trade-off optimal solutions and then in choosing a few preferred solutions for implementation.

\bibliographystyle{tfcad}
\bibliography{references.bib}